\documentclass[10pt,twocolumn,letterpaper]{article}

\usepackage{iccv}

\input CSMacrosV2.tex

\usepackage{epsfig}
\usepackage{graphicx}
\usepackage{amsmath}
\usepackage{amssymb}
\newlength\savewidth\newcommand\shline{\noalign{\global\savewidth\arrayrulewidth
  \global\arrayrulewidth 1pt}\hline\noalign{\global\arrayrulewidth\savewidth}}

\usepackage{bm}

\def\bx{\bm{x}}
\def\by{\bm{y}}

\usepackage{multirow, multicol, eucal}
\usepackage{adjustbox}
\usepackage{caption}
\usepackage{subcaption}
\usepackage{amsfonts}
\def\mT{\mathrm{T}}
\def\mR{\mathbb{R}}

\usepackage{xcolor}
\usepackage{hyperref}
\hypersetup{
  colorlinks,
  citecolor=violet,
  linkcolor=red,
  urlcolor=blue
}

\usepackage[vlined,ruled,linesnumbered]{algorithm2e}

\iccvfinalcopy %

\def\iccvPaperID{4085} %

\ificcvfinal\fi

\begin{document}

\title{FATNN: Fast and Accurate Ternary Neural Networks\thanks{PC and BZ contributed equally. Part of this work was done when all authors were with The University of Adelaide. CS is the corresponding author,  email: \tt chunhua@me.com}
}

\author{
Peng Chen, ~~~~~ Bohan Zhuang, ~~~~~ Chunhua Shen
\\[0.12cm]
Data Science \&  AI,
Monash University, Australia
}

\maketitle
\ificcvfinal\thispagestyle{empty}\fi

\begin{abstract}
Ternary Neural Networks (TNNs) have received much attention due to being potentially orders of magnitude faster in inference, as well as more power efficient, than full-precision counterparts. However, 2 bits are required to encode the ternary representation with only 3 quantization levels leveraged. As a result, conventional TNNs have similar memory consumption and speed compared with the standard 2-bit models, but have worse representational capability. Moreover, there is still a significant gap in accuracy between TNNs and full-precision networks, hampering their deployment to real applications. To tackle these two challenges, in this work, we first show that, under some mild constraints, computational complexity of the ternary inner product can be reduced by $2\times$. Second, to mitigate the performance gap, we elaborately design an implementation-dependent ternary quantization algorithm. The proposed framework is termed Fast and Accurate Ternary Neural Networks (FATNN). Experiments on image classification demonstrate that our FATNN surpasses the state-of-the-arts by a significant margin in accuracy. More importantly, speedup evaluation compared with various precision is analyzed on several platforms, which serves as a strong benchmark for further research.
\end{abstract}

\section{Introduction}  %
\label{sec:introduction}
Equipped with high-performance computing and large-scale datasets, deep convolution neural networks (DCNN) have become a cornerstone for most computer vision tasks.
However, a significant obstacle for deploying DCNN algorithms to mobile/embedded
edge devices with limited
computing resources is the ever growing computation complexity---in order to achieve good accuracy, the models are becoming very heavy.
To tackle this problem, much research effort has been spent on
model compression.
Representative methods include model quantization \cite{zhuang2018towards, zhou2016dorefa}, network pruning \cite{li2017pruning, zhuang2018discrimination} and neural architecture search for lightweight models \cite{zoph2017neural, liu2019darts}.
In this paper, we focus on model quantization, which reduces the model complexity by representing a network with low-precision weights and activations.

Network quantization aims to map the continuous input values within a quantization interval to the corresponding quantization level, and a low-precision quantized value is assigned accordingly.
TNNs in which both the activations and weights are quantized to ternary, are particularly of interest because
most of
the calculations can be realized with bit operations, thus completely
eliminating multiplications. However, there exists two limitations for conventional TNNs. The first limitation is the inefficient implementation of TNNs. Specifically, the ternary representation of $\{-1,0,1\}$ needs 2 bits to encode with one state wasted. As a result, with the conventional bitwise implementation of quantized networks \cite{zhou2016dorefa, zhang2018lq}, the complexity of ternary inner product is the same with the standard 2-bit counterparts. Another limitation is the considerable accuracy drop compared with the full-precision counterparts due to the much more compact capacity.

To handle these drawbacks, we introduce a new framework, termed FATNN, where we co-design the underlying ternary implementation of computation and the quantization algorithm. In terms of implementation, we fully leverage the property of the ternary representation to design a series of bit operations to accomplish the ternary inner product with improved efficiency. In particular, FATNN reduces the computational complexity of TNNs by $2\times$, which solves the existing efficiency bottleneck. In contrast to previous works, nearly no arithmetic operations exist in the proposed implementation. Also, FATNN works efficiently on almost all kinds of devices (such as CPU, GPU, DSP, FPGA and ASIC) with basic bit operation instructions available.
Furthermore, we design the compatible ternary quantization algorithm in accordance to the mild constraints derived from the underlying implementation. %
Early works with learned quantizers either propose to learn the quantized values \cite{zhang2018lq} or seek to learn the quantization intervals \cite{jung2019learning, esser2019learned}. However, most of them assume the uniform quantizer step size, which might still be non-optimal on optimizing network performance.
To make the low-precision discrete values sufficiently fit the statistics of the data distribution, we propose to parameterize the step size of each quantization level and optimize them with the approximate gradient.
Besides, we suggest calibrating the distribution between the skip connection and the low-precision branch to further improve the performance.
The overall approach is usable for quantizing both activations and weights, and works with existing methods for back-propagation and stochastic gradient descent.

Our main contributions are summarized as follows:

\begin{itemize}
\itemsep -0.125cm
    \item We propose a ternary quantization pipeline, in which we co-design the underlying implementation and the quantization algorithm. To our best knowledge, the proposed FATNN is the first solution being applied on general platforms specific to TNNs while previous acceleration methods only target on the dedicated hardware (FPGA or ASIC).

    \item We devise a fast ternary inner product implementation which reduces the complexity of TNNs by $2 \times$ while keeping the bit-operation-compatible merit. We then design a highly accurate ternary quantization algorithm in accordance with the constraints imposed by the implementation. %

    \item We evaluate the execution speed of FATNN and make comparison with other bit configurations
    on various
    platforms. Moreover,
    experiments on image classification task demonstrate the superior performance of our FATNN over a few
    competitive %
    state-of-the-art approaches.
\end{itemize}

\section{Related Work}

Model quantization aims to quantize the weights, activations and even backpropagation gradients into low-precision, to yield highly compact DCNN models compared to their
floating-point counterparts. As a result, most of the multiplication operations in network inference can be replaced by more efficient addition or bitwise operations. In particular, BNNs \cite{rastegari2016xnor, hubara2016binarized, zhao2018bitstream, bethge2018training,bethge2018learning,yang2017bmxnet,tang2017train,guo2017network, liu2018bi}, where both weights and activations are quantized to binary tensors, are reported to have potentially $32\times$ memory compression ratio, and up to $58\times$ speed-up on CPU compared with the full-precision counterparts. However, BNNs still suffer from sizable performance drop issue, hindering them from being widely deployed.
To make a trade-off between accuracy and complexity, researchers also study ternary \cite{li2016ternary,zhu2016trained} and higher-bit quantization \cite{zhou2016dorefa, choi2018pact,esser2019learned,zhuang2019structured, lin2017towards}. In general, quantization algorithms aim at tackling two core challenges. The first challenge is to design accurate quantizers to minimize the information loss. Early works use handcrafted heuristic quantizers \cite{zhou2016dorefa} while later studies propose to adjust the quantizers to the data, basically based on matching the original data distribution \cite{zhou2016dorefa, Cai_2017_CVPR}, minimizing the quantization error \cite{zhang2018lq, Yoojin2018regular} or directly optimizing the quantizer with stochastic gradient descent \cite{choi2018pact, jung2019learning, baskin2018uniq}.
Moreover, another challenge is to approximate gradient of the non-differentiable quantizer.
To solve this problem, most studies focus on improving the training via loss-aware optimization \cite{hou2018loss}, regularization \cite{ding2019regularizing, Yoojin2018regular, bai2019proxquant}, knowledge distillation \cite{zhuang2018towards, mishra2018apprentice}, entropy maximization \cite{park2017weighted, polino2018model} and relaxed optimization \cite{louizos2019relaxed, Yang_2019_CVPR, 2017congADMM, 2018peisongTSQ}.
In addition to the quantization algorithms design, the implementation frameworks and acceleration libraries \cite{ignatov2018ai,chen2018tvm,umuroglu2017finn,jacob2017quantization} are indispensable to expedite the quantization technique to be deployed on energy-efficient edge devices. For example, TBN \cite{Wan_2018_ECCV} focuses on the implementation of ternary activation and binary weight networks. daBNN \cite{zhang2019dabnn} targets at the inference optimization of BNNs on ARM CPU devices.
GXNOR-Net \cite{deng2018gxnor} treats TNNs as a kind of sparse BNNs and propose an acceleration solution on dedicated hardware platforms. RTN \cite{Li2020RTN} leverages extra linear transformation before quantization for better performance of TNNs. Similar with GXNOR-Net, RTN demonstrates the benefit of their implementation on dedicated devices (FPGA and ASIC).
However, there are few works targeting on improving the inference efficiency of TNNs on general purpose computing platforms.
In this paper, we propose to co-design the underlying implementation and the quantization algorithm to achieve ideal efficiency and accuracy simultaneously.

\section{Inference Acceleration}

\subsection{Preliminary}

As the inner product is one of the fundamental operations in convolution neural networks, which consumes most of the execution time, we mainly focus on the acceleration of inner product in this paper. It is worth to firstly review how the inner product between two quantized vectors are computed in previous literature.
For BNNs, in which both the weights and activation are binarized to $\{-1, 1\}$, the inner product between two length-$N$ vectors $\bx, \by \in \{-1,1\}^N$ can be derived using bit-wise operations:
\begin{equation}   \label{eq:bnn}
{\bx} \cdot {\by} = 2 \cdot {\rm{popcount}}({\rm{xnor}}({\bx},{\by})) - N,
\end{equation}
where $\rm{popcount}$ counts the number of bits in a bit vector.
Furthermore, for quantization with more bits, the input vectors can be decomposed with a linear combination of binary bases. For example, a $M$-bit vector $\bx$ can be encoded as $\bx =\sum_{m=0}^{m=M-1}{\bx}_m\cdot 2^m$ where ${\bx}_m \in \{-1, 1\}^N$ (in practical implementation, $-1$ and $1$ are represented by $0$ and $1$ respectively).
Similarly, for another $K$-bit vector $\by$, we have $\by =\sum_{k=0}^{k=K-1}{\by}_k\cdot 2^k$, where ${\by}_k \in \{-1, 1\}^N$. Based on the decomposition, the binary inner product specified in Eq. \eqref{eq:bnn} can be used to compute higher bit inner product.
Generally, the inner product between two quantized vectors can be formulated as
\begin{equation} \label{eq:lqnet}
{\bx} \cdot \by = \sum\limits_{m = 0}^{M - 1} {\sum\limits_{k = 0}^{K - 1} {{\alpha_m}} } {\beta_k}({\bx}_m \odot {\by}_k),
\end{equation}
where $\odot$ is specially used to denote the binary inner product in formulation of Eq. \eqref{eq:bnn}, $\bm{\alpha}\in {\mathbb{R}^M}$ and $\bm{\beta}\in {\mathbb{R}^K}$ are scales to encode $\bx$ and $\by$, respectively.
Specifically, $\alpha_m = 2^m$ and $\beta_k = 2^k$ are used in uniform fixed-point quantization \cite{zhou2016dorefa, lin2016fixed} while $\bm{\alpha}$ and $\bm{\beta}$ become trainable scales for non-uniform quantization \cite{zhang2018lq, baskin2018uniq}. Considerable speedups can be achieved by Eq. \eqref{eq:bnn} and Eq. \eqref{eq:lqnet} because
all the calculation can be realized with bit operations \cite{hubara2016binarized, courbariaux2015binaryconnect, umuroglu2017finn}, thus completely eliminating multiplications.

\subsection{Motivations for Acceleration}

If assuming the computational complexity of BNNs in Eq.~\eqref{eq:bnn} to be $O(N)$, the computational complexity for higher bit quantization in Eq.~\eqref{eq:lqnet} becomes $O(M \cdot K \cdot N)$. We can find that higher bit quantization algorithms acquire better task accuracy at the cost of increased computational complexity. In particular, for the TNN case, 2 bits are required for the data representation of $\{-1,0,1\}$. Thus the computational complexity for ternary inner product is $O(4 N)$, which is the same with the standard 2-bit counterpart, however, with one of the quantization levels wasted (2 bits can express 4 quantization levels at most). As a result, the implementation in Eq. \eqref{eq:lqnet} makes TNNs less appealing to standard 2-bit models in practical.

To fully unleash the potential of TNNs, we further observe that the
binary inner product in Eq.~\eqref{eq:bnn} is the core for acceleration since its multiplication and accumulation are realized by the bit operators  $\rm xnor$ and $\rm popcount$, respectively.
As the input of the inner product in BNNs
is restricted to $\{-1, 1\}$, the multiplication result is also within the set $\{-1, 1\}$
, which we call the ``\textbf{non-overflow}'' property.
The multiplication result can be directly obtained via $\rm xnor$ between the input vectors
and it
owns the
attribute that only two states exist,
thus $\rm popcount$ can be used to realize accumulation by simply counting the number of state ``$1$'' (or state ``$-1$'').
As a result,
Eq.~\eqref{eq:bnn}
enables the same parallelism degree\footnote{
The same parallelism degree indicates the data amount processed is the same per instruction.  More
explanations
are put
in Section S1 in the supplementary file.} for $\rm{xnor}$ and $\rm{popcount}$, with the ALU register fully utilized.
Interestingly, we find the ternary quantized values $\{1, 0, -1\}$ also meets the ``non-overflow'' property.
Thus, it is potential for the TNNs to be executed in the same parallelism degree manner for the multiplication (\textit{i.e.}, $\rm xnor$) and accumulation (\textit{i.e.}, $\rm popcount$) operations. Moreover, we can %
use
this property to design a novel ternary inner product implementation with a reduced complexity of $O(2N)$.

\subsection{Ternary Network Acceleration} \label{sec:implementation}

\begin{table}[t]
\centering
\caption{The correspondence mapping between the quantized data space and the codec space employed in the proposed solution. Both 2'b01 and 2'b10 are taken to represent the ``0'' value. The design of the codec owns the attribute of ``popcount(codec) $=$ data$\,+ 1$''.}
\label{tab:codec}
\begin{tabular}{c | c c c c}

 data  & $-1$ & 0 & 0 & 1 \\
 \shline
 codec & 2'b00 & 2'b01 & 2'b10 & 2'b11 \\

\end{tabular}
\end{table}

We now elaborate the design of the fast ternary inner product implementation.
First, it is worth noting that the ternary values $\{-1, 0, 1\}$ will be represented by the corresponding codec in the practical implementation.
We elaborately design the mapping between the logical level ternary values and their implementation level codec as illustrated in Table \ref{tab:codec}. Interestingly, we observe a property that the number of ``1'' in each value's codec equals the value plus one. Therefore, we can compute the inner product between two ternary vectors $\bx, \by \in \{-1, 0, 1\}^N$ as follows:
\begin{equation} \label{eq:popcount}
\begin{aligned}
{\bx} \cdot {\by} = {\rm{popcount}} ({\rm{TM}}({\bx},{\by})) - N,
\end{aligned}
\end{equation}
where $\rm {TM(\cdot)}$ indicates the ternary multiplication, $N$ is the vector length.
In Eq. \eqref{eq:popcount}, we first compute the inner product in the codec space by ${\rm{popcount}}({\rm{TM}}({\bx},{\by}))$, which consists of pure bit operations. Then we simply subtract $N$ from the result to transform it into the inner product of the logical level ternary vectors.

Second, we illustrate the way to design $\rm {TM(\cdot)}$ with pure bit operations\footnote{We follow the C/C++ grammar in the equations.
For example, $\&$ means ``AND'', $\sim$ indicates ``NOT'', $\wedge$ represents ``XOR''.}.
For easy understanding, the true value table of ternary multiplication is listed in Table \ref{tab:true-table}. Since two bits are required to encode the ternary input value and the ternary multiplication result, there are 16 possibilities with respect to the codec. From Table \ref{tab:true-table}, we observe that most rows in the table, except the bold ones, still follow the rule of $\rm xnor$:
\begin{equation}
\label{eq:tm-1}
{xnor}  = \ \sim (x_c \wedge y_c),
\end{equation}
where $x_c$ and $y_c$ are the codec representation of the element in $\bx$ and $\by$, respectively.
\noindent Therefore, the ternary multiplication can be realized by $\rm xnor$ along with the exception cases fixed.

After reviewing the $\rm xnor$ correct cases and exception cases, we summarize that the exception ones only happen when both of the operands are $0$. Therefore, the ternary multiplication result can be fixed by locating the ``zero operand'' cases and forcing the result to be $0$ (we force the result to be $0$ if ``zero operand'' is detected no matter whether it is the $\rm xnor$ incorrect case or not).
\begin{table}
\centering
\caption{True value table of ternary multiplication.}
\label{tab:true-table}
\small
\begin{tabular}{r r r r r}
\multicolumn{1}{c}{$x_c$} & \multicolumn{1}{c}{$\cdot$} & \multicolumn{1}{c}{$y_c$} & \multicolumn{1}{c}{$=$} & \multicolumn{1}{c}{$z_c$}  \\
\shline
$-1$(2'b00) &  & $-1$(2'b00) &  &  1(2'b11) \\
$-1$(2'b00) &  &  0(2'b01) &  &  0(2'b10) \\
$-1$(2'b00) &  &  0(2'b10) &  &  0(2'b01) \\
$-1$(2'b00) &  &  1(2'b11) &  & $-1$(2'b00) \\

 0(2'b01) &  & $-1$(2'b00) &  &  0(2'b10) \\
 \textbf{0(2'b01)} &  &  \textbf{0(2'b01)} &  &  \textbf{0(2'b01)} \\
 \textbf{0(2'b01)} &  &  \textbf{0(2'b10)} &  &  \textbf{0(2'b01)} \\
 0(2'b01) &  &  1(2'b11) &  &  0(2'b01) \\
 0(2'b10) &  & $-1$(2'b00) &  &  0(2'b01) \\
 \textbf{0(2'b10)} &  &  \textbf{0(2'b01)} &  &  \textbf{0(2'b10)} \\
 \textbf{0(2'b10)} &  &  \textbf{0(2'b10)} &  &  \textbf{0(2'b10)} \\
 0(2'b10) &  &  1(2'b11) &  &  0(2'b10) \\

 1(2'b11) &  & $-1$(2'b00) &  & $-1$(2'b00) \\
 1(2'b11) &  &  0(2'b01) &  &  0(2'b01) \\
 1(2'b11) &  &  0(2'b10) &  &  0(2'b10) \\
 1(2'b11) &  &  1(2'b11) &  &  1(2'b11) \\
\end{tabular}
\end{table}
Overall, our implementation of the ternary multiplication consists of three steps: 1): Obtain the
intermediate result by the $\rm xnor$ operation. 2): Identify the 0 operand. 3): Fix the exceptions and obtain the ternary multiplication result.
The first step is easily achieved by Eq. \eqref{eq:tm-1}.
For the second step, we here introduce an auxiliary variable $auxi$ which is a pre-defined constant in codec $2'b01$. %
In fact, the auxiliary variable $auxi$ indicates a zero value variable (one codec of the $0$ value is $2'b01$), which acts as a mask to fetch specific bits in operands.
Then we propose to identify the $0$ operand using the following bit operations:
\begin{equation}
\label{eq:tm-2}
switch = ((y_c >> 1)\ \&\ auxi)\ |\ ((y_c<<1)\ \&\sim auxi),
\end{equation}
\begin{equation}
\label{eq:tm-3}
mask = switch \wedge y_c,
\end{equation}
\noindent With the shift and mask operations, Eq. \eqref{eq:tm-2} actually results in the exchange of the two sequence bits in the operand.
After that, Eq. \eqref{eq:tm-3} generates the mask information by distinguishing whether the operand is $0$ (in codec $2'b01$ or $2'b10$) or not. Specifically, the $mask$ variable in Eq. \eqref{eq:tm-3} will be $2'b11$ if the 0 operand is detected and $2'b00$ otherwise.
After identifying the $0$ operand, we can easily fix the exceptions and obtain the final ternary multiplication result as:
\begin{equation}
\label{eq:tm-4}
{\rm{TM}}(\cdot) = (mask\ \& \ auxi)\ |\ ((\sim mask)\ \& \ xnor).
\end{equation}
If the $mask$ is $2'b11$ (the 0 operand is detected), then Eq. \eqref{eq:tm-4} reduces to $auxi$ which equals 0.
In contrast, if the $mask$ is $2'b00$, then Eq. \eqref{eq:tm-4} becomes $xnor$, where the correctness can be examined in Table \ref{tab:true-table}.
In practise, we use the weight parameter ($y_c$) to generate $mask$ which is determined after training.

\paragraph{\bf{Remark 1}}
We can derive from Eq.~\eqref{eq:popcount} that the computational complexity of the proposed ternary inner product is $O(2N)$. In other words, FATNN can reduce the computational complexity of TNNs from $O(4N)$ to $O(2N)$, which significantly improves the efficiency (memory consumption is not changed).
Therefore, even though TNNs do not make full use of the 2-bit representational capacity, they fortunately enjoy the faster implementation than the standard 2-bit models.
Besides, our solution can adapt to general purpose computing platforms, such as CPU, GPU and DSP.
Note that, the extra bit operations introduced in Eqs.~\eqref{eq:tm-2}, \eqref{eq:tm-3} and \eqref{eq:tm-4} have negligible runtime overhead for deployment, as these extra bit operations are much faster than the accumulation %
operation in Eq.~\eqref{eq:popcount} \footnote{Refer to the implementation details in Section S1 in the supplementary material.}.
We further provide extensive benchmark results in
the experiment section
to justify our analysis.

\paragraph{Constraints on the algorithm.}
From the formulation discussed above, it can be learned that the designed ternary implementation has certain requirements on the quantization algorithm. The constraints are summarized as follows:
\begin{itemize}
\itemsep 0cm
\item The ternary values for the network are limited to $\{-1, 0, 1\}$. One and only one additional high precision coefficient is allowed to adjust the scale of the quantized values. More than one scale coefficients will break Eq.~\eqref{eq:popcount}. It indicates methods such as TTN \cite{zhu2016trained}, in which two trainable variables $(W_p, W_n)$ are learned, cannot be applied to our method.

\item A special case exists for the activation quantization when the $\rm{ReLU}$ non-linearity is applied, which leads to a non-negative data range. In this situation, we advise to modify the ternary values to $\{0, 1, 2\}$ for activations. The revision does not conflict with the first constraint as in the inference procedure, it results in an additional constant on the output
(%
simply
add a copy of weights on the result which are fixed after training).

\end{itemize}

\section{Constrained Ternary Quantization}
\label{constrained_algorithm}

%

%

%

\subsection{Quantization Function Revisited}
\label{sec:algorithm}

The quantization functions can be categorized into uniform \cite{esser2019learned, zhou2016dorefa} and non-uniform \cite{zhang2018lq, jung2019learning} ones. In particular, LSQ \cite{esser2019learned} proposes to parameterize the uniform step size and achieves state-of-the-art performance. In this section, we show that the principle of parameterized step size can be applied to solving the non-uniform ternary quantization.

Let us first consider the ternary weight quantization.
In particular, we parameterize the three step sizes by introducing two learnable parameters $\{\alpha_1, \alpha_2\}
$.
In this way, the full-precision data is partitioned into three levels with the quantization thresholds $\{-\alpha_1/2, \alpha_2/2\}$, where each step size can be adjusted accordingly during training.
Then we define the weight quantizer $Q^w(p;\alpha_1, \alpha_2)$ for data $p$ parameterized by the scale factors \{$\alpha_1, \alpha_2$\}. Specifically, $Q^w(p;\alpha_1, \alpha_2)$ performs quantization by applying three point-wise operations in order: normalization, saturate and round.

\textbf{Normalization}: Since we do non-uniform quantization, the %
tensor elements are firstly normalized by the scale factors $\alpha_1$ and $\alpha_2$, respectively. This operation aims to map the data from the floating-point domain to the quantized domain.
\textbf{Saturate}: Once normalized%
,
the tensor elements that are out of the range of the quantized domain are clipped accordingly: ${\rm clip}(p;\beta_1, \beta_2) = {\rm min}({\rm max}(p,\beta_1),\beta_2)$, where the scalar clipping limits $\{\beta_1, \beta_2\}$ are independent with the full-precision data range.
\textbf{Round}: We discretize the normalized and tailored tensor elements to nearest integers using bankers rounding denoted by $\left \lfloor \cdot \right \rceil$.
Putting the above point-wise operations together, the weight quantization function can be written as:
\begin{equation}
\label{eq:alg-1}
Q^w(p) = \left\{
\begin{array}{ll}
\left \lfloor {\rm clip}(p / \alpha_1, -1, 0) \right \rceil & {\rm{if}} \;  p < 0 \\
\left \lfloor {\rm clip}(p / \alpha_2, 0, 1) \right \rceil  & \rm{otherwise}
\end{array}
\right.,
\end{equation}
For simplicity, Eq. \eqref{eq:alg-1} can be re-written into a unified form:
\begin{equation}
\label{eq:alg-2}
Q^w(p) =
\lfloor {\rm clip}(p / \alpha_1, -1, 0) \rceil +
\lfloor {\rm clip}(p / \alpha_2, 0, 1)  \rceil.
\end{equation}

Since the bankers rounding is non-differentiable, we use the straight through estimator (STE) \cite{bengio2013estimating} to approximate the gradient through the round function $\left \lfloor \cdot \right \rceil$ as a pass through operation, and differentiating all other operations normally \cite{paszke2017automatic}.

For activation quantization, there exists two cases. On the one hand, if $p$ is in the real domain (e.g., use PReLU or Tanh as non-linearity), the activation quantizer is the same as $Q^w(\cdot)$. On the other hand, if $p$ is in the non-negative domain (applicable to ReLU activations), the quantized values should be adapted accordingly. In this case, the activation quantizer becomes:
\begin{equation}
\label{eq:alg-4}
Q^a(p) =
\lfloor {\rm clip}(p / \alpha_1, 0, 1) \rceil +
\lfloor {\rm clip}(p - \alpha_1) / \alpha_2, 0, 1)  \rceil .
\end{equation}

Note that we learn independent step sizes for weights and activations of each quantized layer. %

\paragraph{\bf{Remark 2}} Above non-uniform step size quantization algorithm can be easily integrated in the general deep learning training framework. It is worth noting that the algorithm meets all the constraints derived in Section \ref{sec:implementation}, which only have requirements on the quantized values while having no requirements on the quantization thresholds.
We further visualize the non-uniform step sizes in Section S2 in the supplementary file to provide more insights.

The proposed algorithm is featured in learning the non-uniform quantization thresholds, which is different from two
categories of methods. Compared with methods using a uniform step size \cite{zhou2016dorefa,choi2018pact,esser2019learned}, our FATNN enjoys more flexible step sizes to better fit the statistics of the data distribution. Moreover, different from non-uniform LQ-Net \cite{zhang2018lq} that updates the quantizer via closed-form approximation and  QIL \cite{jung2019learning} that introduces extra non-linear transformation and hyperparameters with careful tuning during optimization, our FATNN directly parameterizes and updates the quantization intervals by efficient stochastic gradient descent, which shows better performance in Section \ref{exp}.

\subsection{Calibration of Residual Block}

It has been shown in the quantization literature that high-precision skip connections are essential to reduce the accumulated quantization error and ease the difficulty in propagating gradients through a low-precision network due to the non-differentiable quantization function. However, few work pays attention to the influence of quantization on the branch fusion in the residual architecture.

\begin{figure}[tb!]
\centering
\begin{adjustbox}{width=0.6\columnwidth,center}
\includegraphics{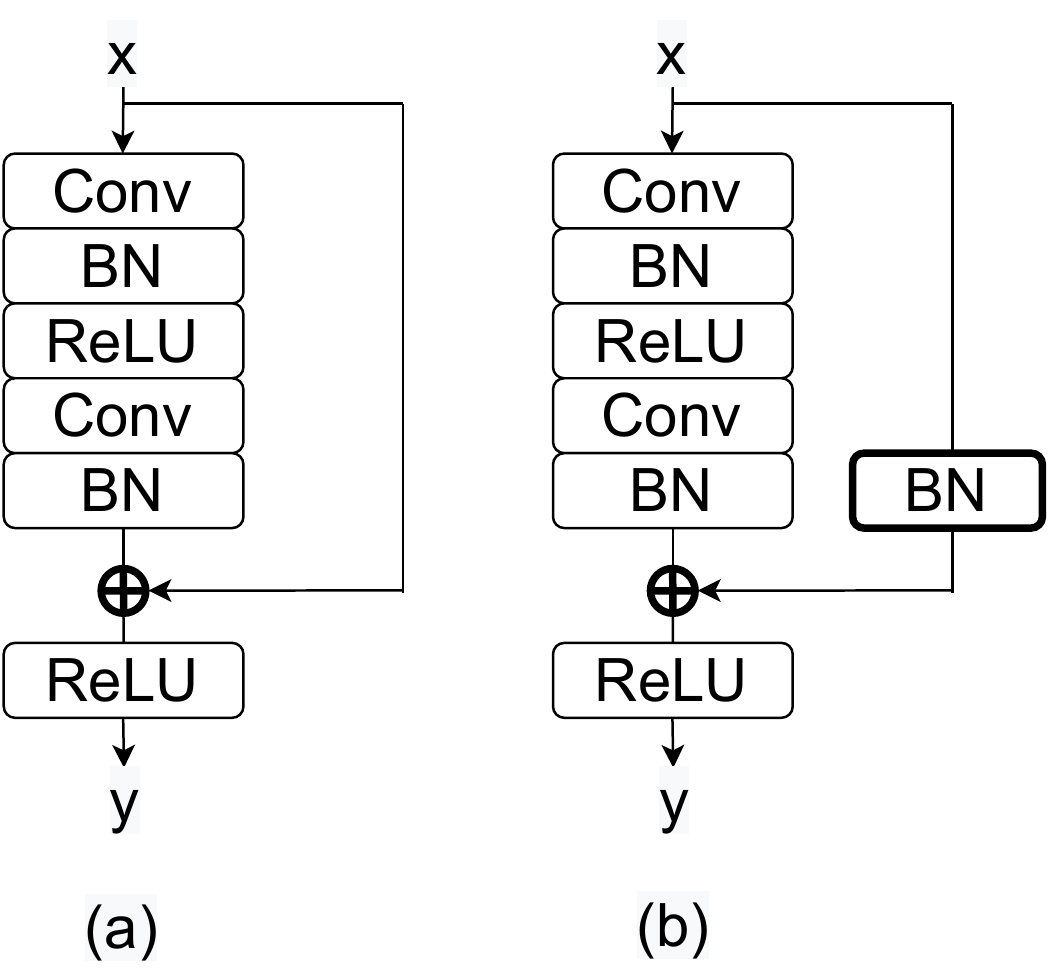}
\end{adjustbox}
\caption{(a). Classical residual block. (b). Calibration between different branches by inserting an extra batch normalization (BN) layer into the identity mapping path.}
\label{fig:branch}
\end{figure}

\begin{figure*}[htp!]
\centering

\begin{subfigure}{.33\textwidth}
  \centering
  \includegraphics[width=1.0\textwidth]{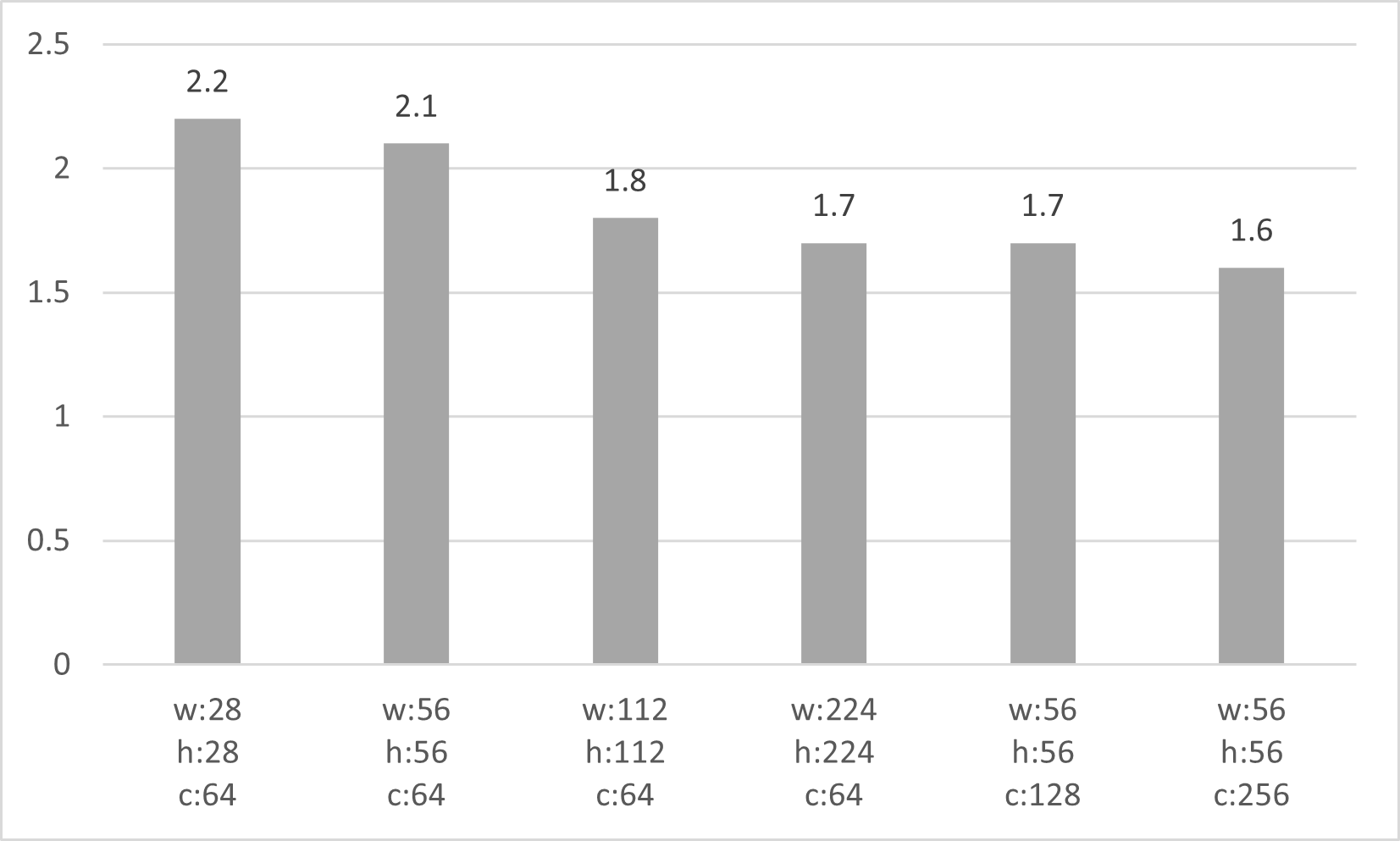}
  \caption{Speedup on Qualcomm 821 GPU}
\label{fig:speedup-1-a}
\end{subfigure}
\begin{subfigure}{.33\textwidth}
  \centering
  \includegraphics[width=1.0\textwidth]{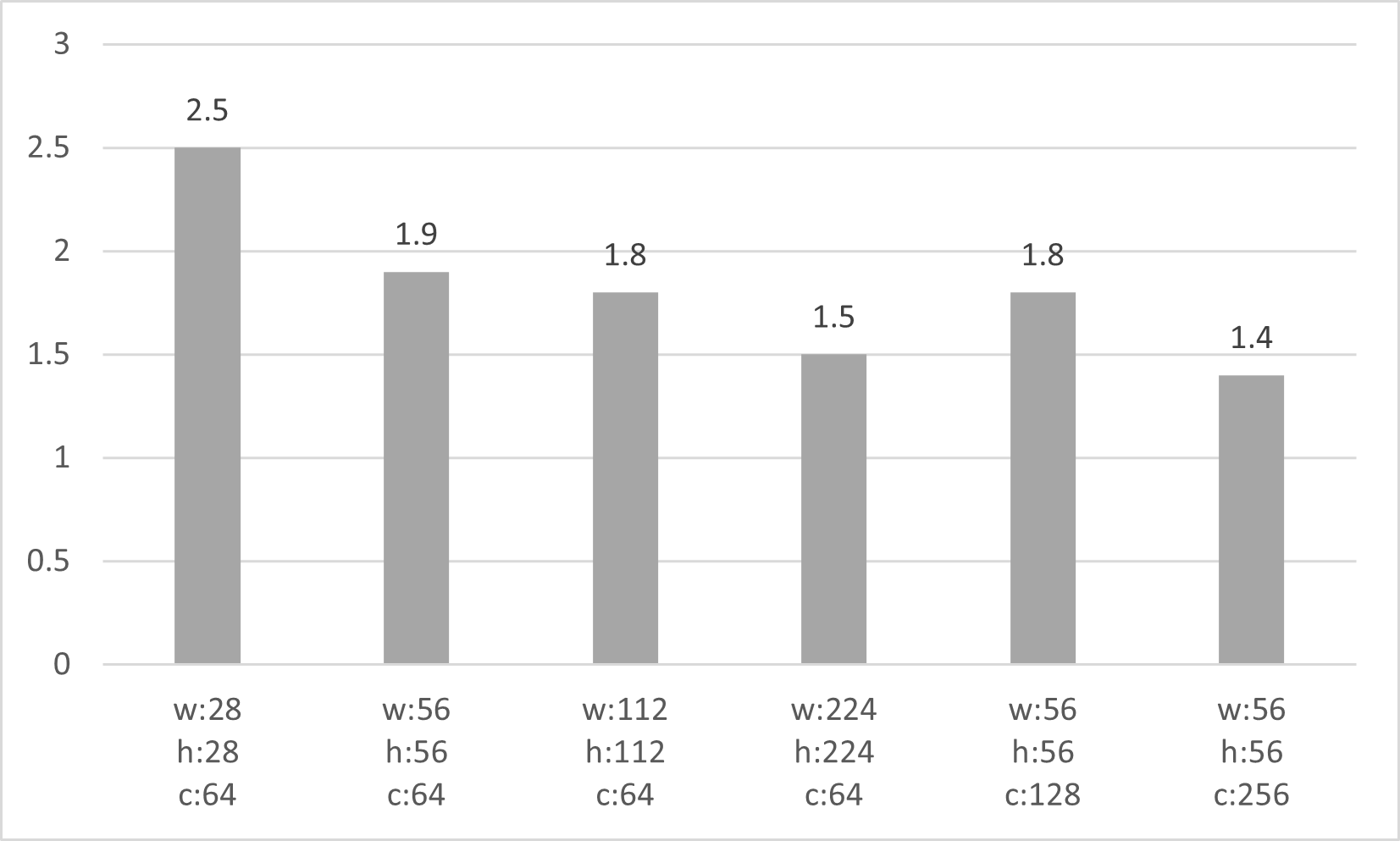}
  \caption{Speedup on Qualcomm 835 GPU}
\label{fig:speedup-1-b}
\end{subfigure}
\begin{subfigure}{.33\textwidth}
  \centering
  \includegraphics[width=1.0\textwidth]{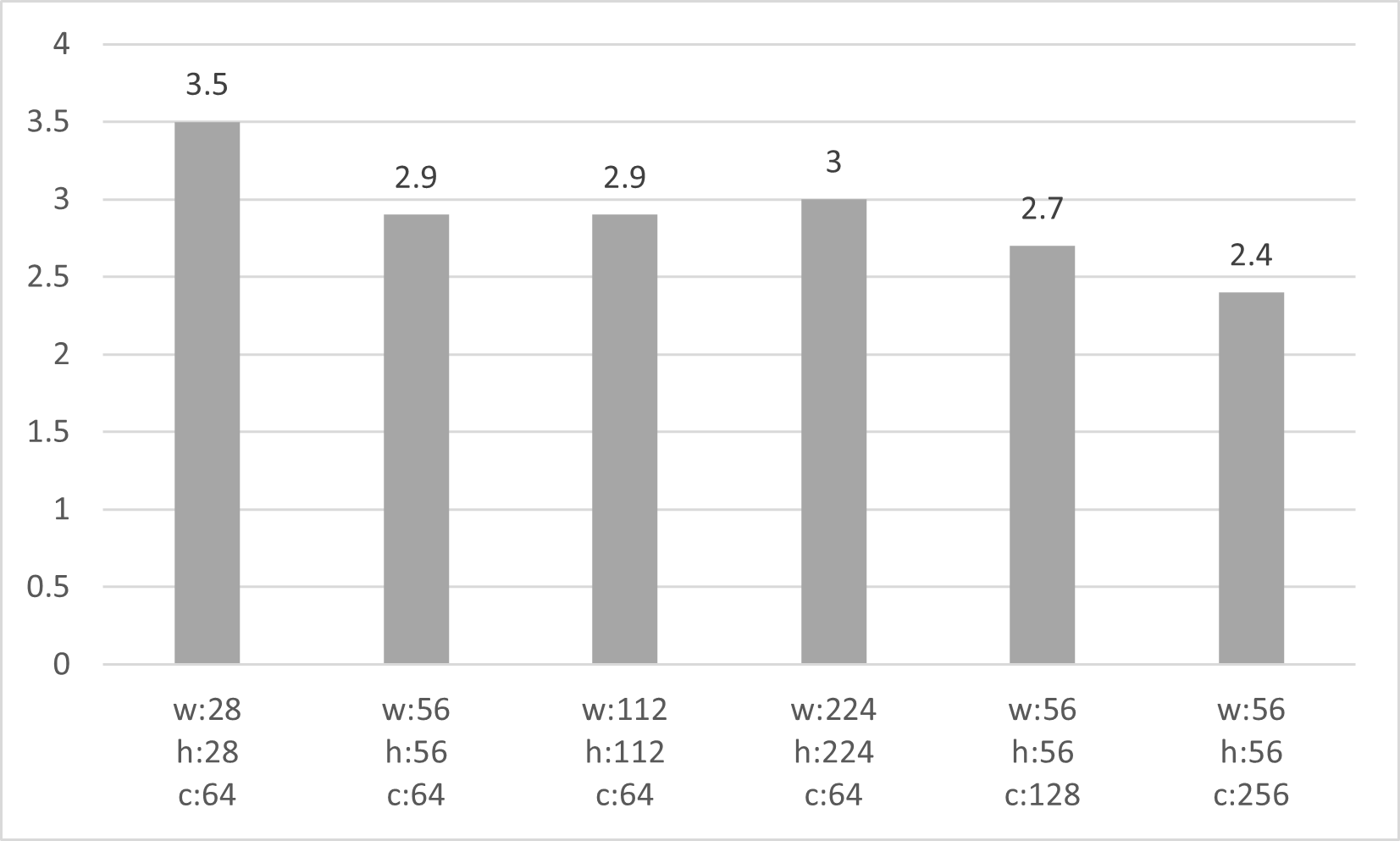}
  \caption{Speedup on Nvidia 2080Ti GPU}
\label{fig:speedup-1-c}
\end{subfigure}

\caption{Layer-wise speedup of ternary networks (`ter') versus 2-bit counterparts (`2-bit') on different platforms.}
\label{fig:speedup-1}

\end{figure*}
As illustrated in Figure \ref{fig:branch} (a), the distributions of the two inputs of the element-wise addition, the non-negative block input $\bx$ (from the preceding residual block) and the output of the batch normalization layer, are different. The large covariate shift between the two inputs results in enlarged variance and value range for the output tensor.
It might not cause a problem for the full-precision network, as
the representational capability with 32-bit precision is enough to encode all the information.
However, it can lead to increased information loss for the low-precision network, because of the deteriorated quantization error when quantizing the enlarged range of value into a set of limited discrete values (\eg, ternary quantization).

To solve this problem, we propose to calibrate the distribution between the skip connection and the low-precision branch.
As illustrated in Figure \ref{fig:branch}(b), the calibration can be done by inserting an extra batch normalization layer in the identity-mapping skip connection path.
The calibration helps to align the two inputs of the element-wise addition, reducing the quantization error when the output tensor is quantized in the succeeding block.
Note that the extra memory consumption is limited as the batch normalization layers only introduce a few parameters (\ie, two scalar parameters for each after training). 
We leverage the BN fusion when the preceding layer is a convolutional layer. Otherwise, extra computation is introduced by the BN layer, typically in the calibration branch. But the runtime overhead is negligible for deployment since the computational cost of BN is orders of magnitude lower than  that of the convolutional layer.  In practice,  we find the execution time of all ``unfused'' BN layers only occupies less than 1\% of that of the whole network.

%

%
\section{Experiments}
\label{exp}

\subsection{Acceleration}  %
\label{sec:acceleration}

In this paper, we propose a fast ternary inner product implementation with $O(2N)$ complexity, which is potentially $2\times$ faster than the implementation in previous works \cite{zhou2016dorefa,zhang2018lq}. To further justify its effectiveness in practice, we develop the acceleration code (in C++ and OpenCL) for binary, ternary and 2-bit convolutions\footnote{We
include
the implementation details in Section S1 of the supplementary material.}.
Measurements of the actual execution time on different devices are conducted. We cover both the embedded-side devices (Qualcomm 821 and Qualcomm 835\footnote{We conduct experiments on Google Pixel and Xiaomi 6 phones which equip with the Qualcomm 821/835 chips respectively. Any platform with the same chip is expected to
produce the similar result.}) and server-side devices (%
Nvidia 2080Ti) to demonstrate the flexibility of our solution. We keep the experimental setting the same across all devices.

\begin{table*}[hbt!]
\centering
\caption{%
Execution time ($\rm ms$) and speedup ratios for overall quantized layers. ``bin'' means binary and ``ter'' is ternary. `Q821' and `Q835' indicate the platform of Qualcomm 821 and Qualcomm 835, respectively. We run 5 times and report the results with mean and standard deviation.}
\label{tab:overall}
{
\small
\begin{tabular}{c | c | c c c | c c c}
 Device & Network & bin & ter & 2-bit & bin vs. ter & bin vs. 2-bit & ter vs. 2-bit \\ \shline
\multirow{2}{*} {Q835} & ResNet-18 & 12.1$\pm$0.2 & 20.1$\pm$0.2 & 43.0$\pm$0.7 & 1.7 & 3.6 & 2.1 \\
  & ResNet-34 & 25.3$\pm$0.3 & 42.1$\pm$0.5  & 87.0$\pm$1.5 & 1.7 & 3.4 & 2.1\\
\hline
\multirow{2}{*} {Q821} & ResNet-18 & 15.7$\pm$0.3 & 25.2$\pm$0.3 & 52.3$\pm$1.0 & 1.6 & 3.3 & 2.1 \\
  & ResNet-34 & 32.3$\pm$0.6 & 51.4$\pm$0.6 & 105.2$\pm$2.1 & 1.6 & 3.3 & 2.0 \\
\end{tabular}}
\end{table*}

\subsubsection{Layer-wise Speedup} \label{sec:layer-wise}

First, we report the layer-wise speedup in Figure \ref{fig:speedup-1}, where convolution layers (kernel size = $3 \times 3$, padding = 1, stride = 1, the same number of input and output channels and batch size = 1) with six different shape configurations are tested. For the first four cases on each platform in Figure \ref{fig:speedup-1}, we fix the channel number to be 64 and increase the resolution from 28 to 224. For the last two cases, we fix the resolution to be 56 and double the channel number from 64 to 256.
We report the relative acceleration ratios between ternary and 2-bit models on Qualcomm 821, 835 and Nvidia 2080Ti platforms in Figure \ref{fig:speedup-1}.
From Figure \ref{fig:speedup-1}, we observe that,
compared with the 2-bit quantization, the proposed ternary quantization is much faster.
Specifically, the ternary convolutional layer is $1.6\times$ to $2.2\times$ faster on the Qualcomm 821 platform, $1.4\times$ to $2.5\times$ faster on the Qualcomm 835 platform and $2.4\times$ to $3.5\times$ faster on the Nvidia 2080Ti platform. The variance of the speedup on different platforms is caused by the macro-architectures of different devices.

\begin{table*}[t]
\centering
\caption{Influence of the extra bit operations in the FATNN implementation. `base' indicates the total layer-wise execution time ($\mu s$) of the ternary convolutional layer. `extra' represents the layer-wise execution time of the extra bit operations described in Eqs. \eqref{eq:tm-2}-\eqref{eq:tm-4}. We report both absolute time and relative ratios. We run 5 times and report the mean results. `Q821', `Q835' and `2080Ti' indicate the platforms of Qualcomm 821, Qualcomm 835 and Nvidia 2080Ti, respectively.
}
\label{tab:two-options}
{
\small
\begin{tabular}{c | r |  c c c c c c}
 Device &  & case1 & case2 & case3 & case4 & case5 & case6 \\
\shline
\multirow{2}{*} {Q821}
& base & 594 & 1190 & 3425 & 12373 & 3631 & 13202 \\
& extra & 61 (10.3\%) & 56 (4.7\%) & 36 (1.1\%) & 21 (0.2\%)& 95 (2.6\%) & 676 (5.1\%)  \\
\hline
\multirow{2}{*} {Q835}
& base & 485 & 1007 & 2847 & 11023 & 2606 & 12122  \\
& extra & 155 (32.0\%)  & 150 (14.9\%) & 471 (16.5\%) & 177 (1.6\%) & 5 (0.2\%) & 268 (2.2\%) \\
\hline
\multirow{2}{*} {2080Ti}
& base & 12  & 15.5 & 29  & 92.5  & 35.5 & 88.5  \\
& extra & 1 (8.3\%) &  1 (6.5\%) & 3 (10.3\%) & 2.5 (2.7\%) & 1.5 (4.2\%) & 1.5 (1.7\%) \\
\end{tabular}}
\end{table*}

\subsubsection{Overall Speedup} \label{sec:overall}

Besides the layer-wise analysis, it is also interesting to get a knowledge of the overall speed for some classical networks. We present the whole execution time of all quantized layers (first and last layers are excluded while non-linear and skip connection layers are included) for ResNet-18 and ResNet-34 on ImageNet in Table \ref{tab:overall}. We set the batch size to 1. From Table \ref{tab:overall}, we can learn that both the 2-bit and ternary models run faster than the theoretical speedup versus the binary counterpart (2 and 4 respectively). Moreover, the speedup of the ternary models compared with the 2-bit ones (last column) again demonstrates that our proposed FATNN can run about $2 \times$ faster than the 2-bit models or conventional TNNs.

\subsubsection{Ablation Study}
In Remark 1, we claim the influence of the extra bit operations is negligible.
To verify the impact, we test the exact execution time of the six cases introduced in Section \ref{sec:layer-wise} on different devices in Table \ref{tab:two-options}.
The total execution time of the implementation described in Section \ref{sec:implementation} is taken as the baseline.
We then measure the cost with respect to the extra bit operations
and list the absolute time and relative ratios in Table \ref{tab:two-options}. From Table \ref{tab:two-options}, we observe that the influence is highly relative to the devices and layer shapes. On Qualcomm 821, Qualcomm 835 and Nvidia 2080Ti, the relative ratios of the extra bit operations range from $0.2\%$ to $10.3\%$, $0.2\%$ to $32.0\%$ and $1.7\%$ to $10.3\%$ on the evaluated cases, respectively.
Besides, the occurrence of the $32.0\%$ ratio happens only on the small input shape ($\rm c=64, w=h=28$) while most of the larger input sizes have relative small ratios. Together with the overall speedup in Section \ref{sec:overall},
we can conclude that the extra bit operations have a small impact on the total execution time.

\subsection{Quantization Accuracy}

We perform experiments on
the ImageNet \cite{russakovsky2015imagenet} dataset.
The ImageNet contains about 1.2 million training and 50K validation images of 1,000 object categories.
To verify the effectiveness of the proposed auxiliary learning strategy, we compare with various representative quantization approaches, including uniform approaches LSQ \cite{esser2019learned} and DoReFa-Net~\cite{zhou2016dorefa} and non-uniform methods LQ-Net~\cite{zhang2018lq}, HWGQ \cite{Cai_2017_CVPR} and QN \cite{Yang_2019_CVPR}. Ternary-oriented quantization methods, such as RTN \cite{Li2020RTN}, TBN \cite{Wan_2018_ECCV} and LSB \cite{Pouransari2020lsb} are also included. %
Comparisons on other datasets, such as CIFAR-10, can be found in Section S4
in the supplementary file.
\subsubsection{Evaluation on ImageNet}

\paragraph{Experimental setup.}
For ImageNet classification, all the images are re-scaled with the shorter edge to be 256. Training images are then randomly cropped into resolution of $224 \times 224$. After that, the images are normalized using the mean and standard deviation. No additional augmentations are performed except the random horizontal flip. Validation images follow a similar procedure except the random crop is replaced with the center crop and no flip is applied.
We conduct experiments on the ResNet models \cite{he2016deep}.
As in previous works \cite{zhou2016dorefa, zhang2018lq}, we do not quantize the first and last layers. Following LSQ \cite{esser2019learned} and IR-net \cite{qin2020forward}, we leverage weight normalization along with the standard batch normalization during training to make the optimization more stable.
If not specially mentioned, the initial learning rate is set to 1e-2 and the cosine annealing decay is employed.
Other default hyper-parameters include: SGD optimizer with a momentum of $0.9$, a weight decay of 2e-5, and a maximum training epoch of 90. The quantization related parameters $\alpha_1$ and $\alpha_2$ are initialized to 1.0 for weights and activations in all quantized layers.
We initialize the quantized network with the pretrained full-precision weights at the beginning of the quantization.

\begin{table*}[hbt!]
\centering
\caption{Accuracy (\%) comparisons between our FATNN and other algorithms on ImageNet. ``A/W'' in the second column indicates the bit configuration for activations and weights respectively. ``ter'' denotes ternary. Results for LSQ are based on our own implementation. Results for algorithms, including TBN, RTN, LSB, LQ-Net, HWGQ, DoReFa-Net and QN, are directly cited from the original papers.}
{\small
\begin{tabular}{l| c | c c | c c | c c }
\multirow{3}{*}{Method} & \multirow{2}{*}{A/W} & \multicolumn{2}{c|}{ResNet-18} & \multicolumn{2}{c|}{ResNet-34} & \multicolumn{2}{c}{ResNet-50}  \\
\cline{3-8}
  & & Top-1 & Top-5 & Top-1 & Top-5 & Top-1 & Top-5  \\
\cline{2-8}
& 32/32 & 69.8 & 89.1 & 73.3 & 91.4 & 76.1 & 92.9  \\
\hline
\textbf{FATNN} (Ours) & ter/ter & \bf{66.0} & \bf{86.4} & \bf{69.8} & \bf{89.1} & \bf{72.4} & \bf{90.6} \\
LSQ \cite{esser2019learned} & ter/ter & 64.7 & 85.6 & 69.0 & 88.8 & 71.2 & 90.1 \\
RTN \cite{Li2020RTN} & ter/ter & 64.5 & - & - & - & - & - \\
\hline
TBN \cite{Wan_2018_ECCV} & ter/1 & 55.6 & 79.0 &  58.2 & 81.0 & - & -  \\
LSB \cite{Pouransari2020lsb} & ter/1 & 62.0 &  83.6 &  - & - & - & - \\
\hline
LSQ \cite{esser2019learned}& 2/1 &  64.9 & 85.8 & 69.1 & 88.8 & 71.0 & 90.0  \\
LQ-Net \cite{zhang2018lq}& 2/1 & 62.6 &  84.3 & 66.6 & 86.9 & 68.7 & 88.4  \\
HWGQ \cite{Cai_2017_CVPR} & 2/1 &  59.6 & 82.2 &  64.3 & 85.7 &  64.6 & 85.9 \\
DoReFa-Net \cite{zhou2016dorefa} & 2/1 &  53.4 & - & - & - & - & -  \\
QN \cite{Yang_2019_CVPR} & 2/1 & 63.4 & 84.9  & - & - & - & - \\
\end{tabular}}
\label{tab:accuracy}
\end{table*}

\paragraph{Performance analysis.}
Firstly, we compare with the current best performed quantization algorithm LSQ, which learns a uniform step size and thus is directly comparable to our FATNN. To make a fair comparison, we implement LSQ under exactly the same training process with FATNN. We report the quantization results in Table \ref{tab:accuracy}.
From Table \ref{tab:accuracy}, we observe steady Top-1 accuracy improvement of our FATNN over LSQ on all comparing architectures in the ternary case. This result strongly justifies the effectiveness of the proposed distribution-aware quantization strategy.
To be emphasized, our FATNN boosts the implementation speed of LSQ by a factor of two according to Table \ref{tab:overall} while still achieving the state-of-the-art accuracy.

Secondly, to make a comprehensive comparison, we also include the results of the ``2-bit activation and binary weight'' networks which have a similar computational complexity $O(2N)$ with our FATNN.
We can learn from Table \ref{tab:accuracy} that our FATNN is able to achieve steady accuracy gain over the non-uniform quantization algorithms, such as LQ-Net, HWGQ and QN. For example, FATNN outperforms LQ-Net by 3.4\% on the Top-1 accuracy on ResNet-18. This justifies the superiority of our gradient-based approach to learn the step sizes. Moreover, compared to the uniform quantization algorithms, for example DoReFa-Net and LSQ, we also achieve the best performance on various architectures. This further shows that the employed algorithm can better fit the statistics of the data distribution to reduce the quantization error.

\subsubsection{Ablation Study}

In this section, we further investigate the effect of the distribution-aware quantization strategies proposed in Section \ref{constrained_algorithm}, including the non-uniform quantization functions and the calibration of residual block, to the final performance respectively.
We conduct the experiments with ResNet-18 on ImageNet and report the results in Table \ref{tab:distribution}.
We observe that the branch calibration strategy deteriorates the performance in the full-precision setting, leading to $0.2\%$ Top-1 accuracy drop. In contrast, the calibration strategy brings $0.6\%$ performance improvement to our ternary FATNN in terms of the Top-1 accuracy.
It empirically justifies that calibrating the residual block is particularly designed for the low-precision network.
Moreover, the non-uniform step size quantization of FATNN further brings $0.7\%$ Top-1 accuracy increase, which justifies that the parameterized quantization thresholds can fit the statistics of the data distribution effectively.

\begin{table}[hbt!]
\centering
\caption{Impact of the distribution-aware ternary quantization algorithms with ResNet-18 on ImageNet. `N' indicates the non-uniform step size quantization algorithm. `C' implies the distribution calibration between the skip connection and the low-precision branch.}
{
\small
\begin{tabular}{l | c | c  c | c c }
{Method} & {A/W} & N & C & Top-1 & Top-5  \\
\hline
& 32/32 &  &  & 69.8 & 89.1 \\
\cline{2-6}
& 32/32 & & \checkmark & 69.6 & 88.9 \\
\hline
LSQ \cite{esser2019learned} (baseline) & ter/ter & & & 64.7 & 85.6 \\
\hline
\multirow{2}{*}{\textbf{FATNN} (Ours)} & ter/ter & \checkmark &  & 65.4 & 86.2 \\
 & ter/ter & \checkmark & \checkmark & \bf{66.0} & \bf{86.4} \\
\end{tabular}}
\label{tab:distribution}
\end{table}

\section{Conclusion}

In this paper, we have proposed a fast ternary neural network, named FATNN. Specifically, we emphasize that the underlying implementation and the
quantization algorithm are highly correlated and should be co-designed. From the implementation perspective, we exploit the ``non-overflow'' property to design a novel ternary inner product with fully bit operations. As a result, our FATNN can achieve $2\times$ less complexity than the conventional TNNs. Moreover, we have designed an efficient quantization algorithm in accordance with the constraints of the implementation.
Extensive experiments %
demonstrate
that FATNN improves the previous TNNs in %
both execution time and quantization accuracy.
Thus,
we advocate to rethink the value of TNNs and %
hope
that FATNN %
can
serve as a strong benchmark for further research.

{\small
\bibliographystyle{ieee_fullname}
\bibliography{egbib}

\begin{thebibliography}{10}\itemsep=-1pt

\bibitem{bai2019proxquant}
Yu Bai, Yu-Xiang Wang, and Edo Liberty.
\newblock Proxquant: Quantized neural networks via proximal operators.
\newblock In {\em Proc. Int. Conf. Learn. Repren.}, 2019.

\bibitem{baskin2018uniq}
Chaim Baskin, Eli Schwartz, Evgenii Zheltonozhskii, Natan Liss, Raja Giryes,
  Alex~M Bronstein, and Avi Mendelson.
\newblock Uniq: Uniform noise injection for non-uniform quantization of neural
  networks.
\newblock {\em arXiv preprint arXiv:1804.10969}, 2018.

\bibitem{bengio2013estimating}
Yoshua Bengio, Nicholas L{\'e}onard, and Aaron Courville.
\newblock Estimating or propagating gradients through stochastic neurons for
  conditional computation.
\newblock {\em arXiv preprint arXiv:1308.3432}, 2013.

\bibitem{bethge2018training}
Joseph Bethge, Marvin Bornstein, Adrian Loy, Haojin Yang, and Christoph Meinel.
\newblock Training competitive binary neural networks from scratch.
\newblock {\em arXiv preprint arXiv:1812.01965}, 2018.

\bibitem{bethge2018learning}
Joseph Bethge, Haojin Yang, Christian Bartz, and Christoph Meinel.
\newblock Learning to train a binary neural network.
\newblock {\em arXiv preprint arXiv:1809.10463}, 2018.

\bibitem{Cai_2017_CVPR}
Zhaowei Cai, Xiaodong He, Jian Sun, and Nuno Vasconcelos.
\newblock Deep learning with low precision by half-wave gaussian quantization.
\newblock In {\em Proc. IEEE Conf. Comp. Vis. Patt. Recogn.}, pages 5918--5926,
  2017.

\bibitem{chen2018tvm}
Tianqi Chen, Thierry Moreau, Ziheng Jiang, Lianmin Zheng, Eddie Yan, Haichen
  Shen, Meghan Cowan, Leyuan Wang, Yuwei Hu, Luis Ceze, et~al.
\newblock {TVM}: An automated end-to-end optimizing compiler for deep learning.
\newblock In {\em USENIX Symp. Operating Systems Design \& Implementation},
  pages 578--594, 2018.

\bibitem{choi2018pact}
Jungwook Choi, Zhuo Wang, Swagath Venkataramani, Pierce I-Jen Chuang,
  Vijayalakshmi Srinivasan, and Kailash Gopalakrishnan.
\newblock Pact: Parameterized clipping activation for quantized neural
  networks.
\newblock {\em arXiv preprint arXiv:1805.06085}, 2018.

\bibitem{Yoojin2018regular}
Yoojin Choi, Mostafa El-Khamy, and Jungwon Lee.
\newblock Learning low precision deep neural networks through regularization.
\newblock {\em ArXiv}, abs/1809.00095, 2018.

\bibitem{courbariaux2015binaryconnect}
Matthieu Courbariaux, Yoshua Bengio, and Jean-Pierre David.
\newblock Binaryconnect: Training deep neural networks with binary weights
  during propagations.
\newblock In {\em Proc. Adv. Neural Inf. Process. Syst.}, pages 3123--3131,
  2015.

\bibitem{deng2018gxnor}
Lei Deng, Peng Jiao, Jing Pei, Zhenzhi Wu, and Guoqi Li.
\newblock Gxnor-net: Training deep neural networks with ternary weights and
  activations without full-precision memory under a unified discretization
  framework.
\newblock {\em Neural Networks}, 100:49--58, 2018.

\bibitem{ding2019regularizing}
Ruizhou Ding, Ting-Wu Chin, Zeye Liu, and Diana Marculescu.
\newblock Regularizing activation distribution for training binarized deep
  networks.
\newblock In {\em Proc. IEEE Conf. Comp. Vis. Patt. Recogn.}, pages
  11408--11417, 2019.

\bibitem{esser2019learned}
Steven~K Esser, Jeffrey~L McKinstry, Deepika Bablani, Rathinakumar Appuswamy,
  and Dharmendra~S Modha.
\newblock Learned step size quantization.
\newblock In {\em Proc. Int. Conf. Learn. Repren.}, 2020.

\bibitem{guo2017network}
Yiwen Guo, Anbang Yao, Hao Zhao, and Yurong Chen.
\newblock Network sketching: Exploiting binary structure in deep cnns.
\newblock In {\em Proc. IEEE Conf. Comp. Vis. Patt. Recogn.}, pages 5955--5963,
  2017.

\bibitem{he2016deep}
Kaiming He, Xiangyu Zhang, Shaoqing Ren, and Jian Sun.
\newblock Deep residual learning for image recognition.
\newblock In {\em Proc. IEEE Conf. Comp. Vis. Patt. Recogn.}, pages 770--778,
  2016.

\bibitem{hou2018loss}
Lu Hou and James~T Kwok.
\newblock Loss-aware weight quantization of deep networks.
\newblock In {\em Proc. Int. Conf. Learn. Repren.}, 2018.

\bibitem{hubara2016binarized}
Itay Hubara, Matthieu Courbariaux, Daniel Soudry, Ran El-Yaniv, and Yoshua
  Bengio.
\newblock Binarized neural networks.
\newblock In {\em Proc. Adv. Neural Inf. Process. Syst.}, pages 4107--4115,
  2016.

\bibitem{ignatov2018ai}
Andrey Ignatov, Radu Timofte, William Chou, Ke Wang, Max Wu, Tim Hartley, and
  Luc Van~Gool.
\newblock Ai benchmark: Running deep neural networks on android smartphones.
\newblock In {\em Proc. Eur. Conf. Comp. Vis.}, 2018.

\bibitem{jacob2017quantization}
Benoit Jacob, Skirmantas Kligys, Bo Chen, Menglong Zhu, Matthew Tang, Andrew
  Howard, Hartwig Adam, and Dmitry Kalenichenko.
\newblock Quantization and training of neural networks for efficient
  integer-arithmetic-only inference.
\newblock In {\em Proc. IEEE Conf. Comp. Vis. Patt. Recogn.}, 2018.

\bibitem{jung2019learning}
Sangil Jung, Changyong Son, Seohyung Lee, Jinwoo Son, Jae-Joon Han, Youngjun
  Kwak, Sung~Ju Hwang, and Changkyu Choi.
\newblock Learning to quantize deep networks by optimizing quantization
  intervals with task loss.
\newblock In {\em Proc. IEEE Conf. Comp. Vis. Patt. Recogn.}, pages 4350--4359,
  2019.

\bibitem{krizhevsky2009learning}
Alex Krizhevsky and Geoffrey Hinton.
\newblock Learning multiple layers of features from tiny images.
\newblock 2009.

\bibitem{2017congADMM}
Cong Leng, Hao Li, Shenghuo Zhu, and Rong Jin.
\newblock Extremely low bit neural network: Squeeze the last bit out with
  {ADMM}.
\newblock {\em CoRR}, abs/1707.09870, 2017.

\bibitem{li2016ternary}
Fengfu Li, Bo Zhang, and Bin Liu.
\newblock Ternary weight networks.
\newblock {\em arXiv preprint arXiv:1605.04711}, 2016.

\bibitem{li2017pruning}
Hao Li, Asim Kadav, Igor Durdanovic, Hanan Samet, and Hans~Peter Graf.
\newblock Pruning filters for efficient convnets.
\newblock {\em arXiv preprint arXiv:1608.08710}, 2017.

\bibitem{Li2020RTN}
Yuhang Li, Xin Dong, Sai~Qian Zhang, Haoli Bai, Yuanpeng Chen, and Wei Wang.
\newblock Rtn: Reparameterized ternary network.
\newblock In {\em AAAI}, 2020.

\bibitem{lin2016fixed}
Darryl Lin, Sachin Talathi, and Sreekanth Annapureddy.
\newblock Fixed point quantization of deep convolutional networks.
\newblock In {\em Proc. Int. Conf. Mach. Learn.}, pages 2849--2858, 2016.

\bibitem{Lin2014NetworkIN}
Min Lin, Qiang Chen, and Shuicheng Yan.
\newblock Network in network.
\newblock {\em arXiv preprint arXiv:1312.4400}, 2013.

\bibitem{lin2017towards}
Xiaofan Lin, Cong Zhao, and Wei Pan.
\newblock Towards accurate binary convolutional neural network.
\newblock In {\em Proc. Adv. Neural Inf. Process. Syst.}, pages 344--352, 2017.

\bibitem{liu2019darts}
Hanxiao Liu, Karen Simonyan, and Yiming Yang.
\newblock Darts: Differentiable architecture search.
\newblock In {\em Proc. Int. Conf. Learn. Repren.}, 2019.

\bibitem{liu2018bi}
Zechun Liu, Wenhan Luo, Baoyuan Wu, Xin Yang, Wei Liu, and Kwang-Ting Cheng.
\newblock Bi-real net: Binarizing deep network towards real-network
  performance.
\newblock {\em arXiv preprint arXiv:1811.01335}, 2018.

\bibitem{louizos2019relaxed}
Christos Louizos, Matthias Reisser, Tijmen Blankevoort, Efstratios Gavves, and
  Max Welling.
\newblock Relaxed quantization for discretized neural networks.
\newblock In {\em Proc. Int. Conf. Learn. Repren.}, 2019.

\bibitem{mishra2018apprentice}
Asit Mishra and Debbie Marr.
\newblock Apprentice: Using knowledge distillation techniques to improve
  low-precision network accuracy.
\newblock In {\em Proc. Int. Conf. Learn. Repren.}, 2018.

\bibitem{park2017weighted}
Eunhyeok Park, Junwhan Ahn, and Sungjoo Yoo.
\newblock Weighted-entropy-based quantization for deep neural networks.
\newblock In {\em Proc. IEEE Conf. Comp. Vis. Patt. Recogn.}, pages 5456--5464,
  2017.

\bibitem{paszke2017automatic}
Adam Paszke, Sam Gross, Soumith Chintala, Gregory Chanan, Edward Yang, Zachary
  DeVito, Zeming Lin, Alban Desmaison, Luca Antiga, and Adam Lerer.
\newblock Automatic differentiation in pytorch.
\newblock In {\em Proc. Adv. Neural Inf. Process. Syst. Workshops}, 2017.

\bibitem{polino2018model}
Antonio Polino, Razvan Pascanu, and Dan Alistarh.
\newblock Model compression via distillation and quantization.
\newblock In {\em Proc. Int. Conf. Learn. Repren.}, 2018.

\bibitem{Pouransari2020lsb}
Hadi Pouransari, Zhucheng Tu, and Oncel Tuzel.
\newblock Least squares binary quantization of neural networks.
\newblock In {\em Proc. IEEE Conf. Comp. Vis. Patt. Recogn. Workshops}, 2020.

\bibitem{qin2020forward}
Haotong Qin, Ruihao Gong, Xianglong Liu, Mingzhu Shen, Ziran Wei, Fengwei Yu,
  and Jingkuan Song.
\newblock Forward and backward information retention for accurate binary neural
  networks.
\newblock In {\em Proc. IEEE Conf. Comp. Vis. Patt. Recogn.}, 2020.

\bibitem{rastegari2016xnor}
Mohammad Rastegari, Vicente Ordonez, Joseph Redmon, and Ali Farhadi.
\newblock Xnor-net: Imagenet classification using binary convolutional neural
  networks.
\newblock In {\em Proc. Eur. Conf. Comp. Vis.}, pages 525--542, 2016.

\bibitem{russakovsky2015imagenet}
Olga Russakovsky, Jia Deng, Hao Su, Jonathan Krause, Sanjeev Satheesh, Sean Ma,
  Zhiheng Huang, Andrej Karpathy, Aditya Khosla, Michael Bernstein, et~al.
\newblock Imagenet large scale visual recognition challenge.
\newblock {\em Int. J. Comp. Vis.}, 115(3):211--252, 2015.

\bibitem{simonyan2014very}
Karen Simonyan and Andrew Zisserman.
\newblock {Very deep convolutional networks for large-scale image recognition}.
\newblock In {\em Proc. Int. Conf. Learn. Repren.}, 2015.

\bibitem{tang2017train}
Wei Tang, Gang Hua, and Liang Wang.
\newblock How to train a compact binary neural network with high accuracy?
\newblock In {\em Proc. AAAI Conf. on Arti. Intel.}, pages 2625--2631, 2017.

\bibitem{umuroglu2017finn}
Yaman Umuroglu, Nicholas Fraser, Giulio Gambardella, Michaela Blott, Philip
  Leong, Magnus Jahre, and Kees Vissers.
\newblock Finn: A framework for fast, scalable binarized neural network
  inference.
\newblock In {\em Proc. ACM/SIGDA Int. Symp. Field-Programmable Gate Arrays},
  pages 65--74. ACM, 2017.

\bibitem{Wan_2018_ECCV}
Diwen Wan, Fumin Shen, Li Liu, Fan Zhu, Jie Qin, Ling Shao, and Heng Tao~Shen.
\newblock Tbn: Convolutional neural network with ternary inputs and binary
  weights.
\newblock In {\em Proc. Eur. Conf. Comp. Vis.}, September 2018.

\bibitem{2018peisongTSQ}
Peisong Wang, Qinghao Hu, Yifan Zhang, Chunjie Zhang, Yang Liu, and Jian Cheng.
\newblock Two-step quantization for low-bit neural networks.
\newblock In {\em Proc. IEEE Conf. Comp. Vis. Patt. Recogn.}, pages 4376--4384,
  2018.

\bibitem{xygkis2018winograd}
Athanasios Xygkis, Lazaros Papadopoulos, David Moloney, Dimitrios Soudris, and
  Sofiane Yous.
\newblock Efficient winograd-based convolution kernel implementation on edge
  devices.
\newblock In {\em Proc. 55th Annual Design Automation Conf.}, DAC '18, pages
  136:1--136:6, New York, NY, USA, 2018.

\bibitem{yang2017bmxnet}
Haojin Yang, Martin Fritzsche, Christian Bartz, and Christoph Meinel.
\newblock Bmxnet: An open-source binary neural network implementation based on
  mxnet.
\newblock In {\em Proc. of the ACM Int. Conf. on Multimedia.}, pages
  1209--1212. ACM, 2017.

\bibitem{Yang_2019_CVPR}
Jiwei Yang, Xu Shen, Jun Xing, Xinmei Tian, Houqiang Li, Bing Deng, Jianqiang
  Huang, and Xian-sheng Hua.
\newblock Quantization networks.
\newblock In {\em Proc. IEEE Conf. Comp. Vis. Patt. Recogn.}, June 2019.

\bibitem{zhang2018lq}
Dongqing Zhang, Jiaolong Yang, Dongqiangzi Ye, and Gang Hua.
\newblock Lq-nets: Learned quantization for highly accurate and compact deep
  neural networks.
\newblock In {\em Proc. Eur. Conf. Comp. Vis.}, 2018.

\bibitem{zhang2019dabnn}
Jianhao Zhang, Yingwei Pan, Ting Yao, He Zhao, and Tao Mei.
\newblock dabnn: A super fast inference framework for binary neural networks on
  arm devices.
\newblock In {\em Proc. ACM Int. Conf. Multimedia}, pages 2272--2275, 2019.

\bibitem{zhao2018bitstream}
Tianli Zhao, Xiangyu He, Jian Cheng, and Jing Hu.
\newblock Bitstream: Efficient computing architecture for real-time low-power
  inference of binary neural networks on {CPU}s.
\newblock In {\em Proc. ACM Int. Conf. Multimedia}, pages 1545--1552, 2018.

\bibitem{zhou2016dorefa}
Shuchang Zhou, Yuxin Wu, Zekun Ni, Xinyu Zhou, He Wen, and Yuheng Zou.
\newblock Dorefa-net: Training low bitwidth convolutional neural networks with
  low bitwidth gradients.
\newblock {\em arXiv preprint arXiv:1606.06160}, 2016.

\bibitem{zhu2016trained}
Chenzhuo Zhu, Song Han, Huizi Mao, and William~J Dally.
\newblock Trained ternary quantization.
\newblock In {\em Proc. Int. Conf. Learn. Repren.}, 2017.

\bibitem{zhuang2018towards}
Bohan Zhuang, Chunhua Shen, Mingkui Tan, Lingqiao Liu, and Ian Reid.
\newblock Towards effective low-bitwidth convolutional neural networks.
\newblock In {\em Proc. IEEE Conf. Comp. Vis. Patt. Recogn.}, 2018.

\bibitem{zhuang2019structured}
Bohan Zhuang, Chunhua Shen, Mingkui Tan, Lingqiao Liu, and Ian Reid.
\newblock Structured binary neural network for accurate image classification
  and semantic segmentation.
\newblock In {\em Proc. IEEE Conf. Comp. Vis. Patt. Recogn.}, 2019.

\bibitem{zhuang2018discrimination}
Zhuangwei Zhuang, Mingkui Tan, Bohan Zhuang, Jing Liu, Yong Guo, Qingyao Wu,
  Junzhou Huang, and Jinhui Zhu.
\newblock Discrimination-aware channel pruning for deep neural networks.
\newblock In {\em Proc. Adv. Neural Inf. Process. Syst.}, pages 875--886, 2018.

\bibitem{zoph2017neural}
Barret Zoph and Quoc~V Le.
\newblock Neural architecture search with reinforcement learning.
\newblock In {\em Proc. Int. Conf. Learn. Repren.}, 2017.

\end{thebibliography}
}

\def\mT{\mathrm{T}}
\def\mR{\mathbb{R}}

\setcounter{section}{0}

\renewcommand{\thesection}{S\arabic{section}}

\section*{Supplementary Material}

\renewcommand\thefigure{S\arabic{figure}}
\renewcommand{\thetable}{S\arabic{table}}

\section{Acceleration Details}

As a fundamental operation in convolutional neural networks, the vector inner product is one of the core components
in acceleration.
In this section,
we first elaborate more about the ``non-overflow'' property and the same parallelism degree as well as their importance to the fast implementation.
Then we present the detailed implementation of the proposed fast ternary inner product and introduce its application in the convolutional and fully-connected layers.

\noindent{\bf{Non-overflow property:}} If the range of input quantized vectors and the multiplication result in the inner product keeps the same, we term it ``non-overflow'' property. The multiplication result here, indicates the intermediate result of multiplication between the two input quantized vectors.  Binary quantized values $\{1, -1\}$ and ternary quantized values $\{1, 0, -1\}$ are examples. We attribute the fast implementation of our FATNN to the ``non-overflow'' property because it enables the same parallelism degree for the
multiplication (i.e., $\rm xnor$) and accumulation (i.e., $\rm popcount$) operations.
Specifically, for the BNNs case, 8 full-precision values can be packed into one byte. When $\rm xnor$ is employed for the multiplication, the parallelism degree is 8. Moreover, $\rm popcount$ also accumulates 8 data at the same time (same parallelism degree with the $\rm xnor$ operation). For the TNNs case, only 4 full-precision values can be packed into one byte. Thus, the parallelism degree of our $\rm{TM(\cdot)}$ in Eq. (3) is 4. Interestingly, the $\rm popcount$ operation can also accumulate the 4 data simultaneously. However, if standard 2-bit quantization values $\{0, 1, 2, 3\}$ are leveraged, the multiplication by combination of bit-wise operators (such as $\rm xnor$) can own the parallelism degree of 4 if packed in byte. However, 4 bits are required to encode the multiplication result. Thus, the parallelism degree for the accumulation procedure is 2 at most (less than 4). Consequently, the computational efficiency will be halved.
Based on the analysis, it can be learnt that the ``non-overflow'' property is an important attribute to enable the fast implementation of ternary and binary networks.

\begin{algorithm}
\SetAlgoLined
\KwIn
{
(1): Full-precision weight vector $\bm{w} \in \mathbb{R}^N$ and activation vector $\bm{a} \in \mathbb{R}^N$. (2): Pre-allocated temporary buffer $\hat{\bm{w}}$ and $\hat{\bm{a}}$ with unsigned char type in the length of $N/4$. (3): Quantization parameters $\alpha_1^w, \alpha_2^w$ and $\alpha_1^a, \alpha_2^a$ which are used to parameterize the step sizes for weights and activations, respectively.
}
\KwOut{The ternary inner product result $z$.}
    Step (1): data packing;\\
    \For{$i\gets0$ \KwTo $\frac{N-1}{4}$}{
        Pack 4 values of $\bm{a}[4i:4i+3]$ into one unsigned char and store it in $\hat{\bm{a}}[i]$;\\
        Pack 4 values of $\bm{w}[4i:4i+3]$ into one unsigned char and store it in $\hat{\bm{w}}[i]$;\\
    }
    Step (2): ternary inner product;\\
    $acc$ = 0;\\
    \For{$i\gets0$ \KwTo $\frac{N-1}{4}$}{
        $acc$ += {\rm{popcount}}({\rm{TM}}($\hat{\bm{w}}[i]$, $\hat{\bm{a}}[i]$)); \\
    }
    $z = acc - N$
\caption{Fast Ternary Inner Product} \label{alg:inference}
\end{algorithm}

As explained in the Section 3, we design a fast implementation for TNNs by exploiting the ``non-overflow'' property.
Algorithm \ref{alg:inference} summarizes the inference flow of the proposed fast ternary inner product. In Algorithm \ref{alg:inference}, lines $2\sim 5$ quantize the full-precision input vector into the codec of the quantized values. As 2 bits are required to encode one ternary value, 4 full-precision values can be packed into one byte. It is worth noting that it is also possible to pack the full-precision data into other data structures. For example, 8 full-precision data can be packed into $\rm short$ type or 16 full-precision data can be packed into 32-bit $\rm int$ type. More specifically, during the packing, each full-precision data is compared with the corresponding quantization thresholds characterized by \{$\alpha_1$, $\alpha_2$\} and assigned to the corresponding codec value. After that, based on
Eqs. (5), (6), (7), the accumulation is performed as lines $8 \sim 10$ in Algorithm \ref{alg:inference}. Finally the logic level inner product result is obtained according to Eq. (3) in line 11.

Following the implementation of the fast ternary inner product, the convolutional layer can be realized by first expanding the input activation into a matrix ($\rm im2col$) and then conducting the matrix multiplication ($\rm gemm$). To enhance the efficiency, the data packing in Algorithm \ref{alg:inference} is integrated into the $\rm im2col$ operation. After that, the matrix multiplication is realized based on the inner product step in Algorithm \ref{alg:inference}. Tricks, such as winograd \cite{xygkis2018winograd}, are commonly employed in the $\rm gemm$ operation, however we do not integrate these tricks for simplicity.
The fully-connected layer is similar to the implementation of the convolutional layer, which can be regarded as a special case of the latter with kernel size being equal to the feature map resolution.
Other operations, such as the ReLU non-linearity and skip connection layers, can be fused in the $\rm im2col$ procedure. Besides, we fuse the batch normalization layers into the corresponding convolutional or fully-connected layers. %

\section{Visualization of Non-uniform Step Sizes}

\begin{figure}[tb!]
\centering
\begin{adjustbox}{width=0.765\columnwidth,center}
\includegraphics{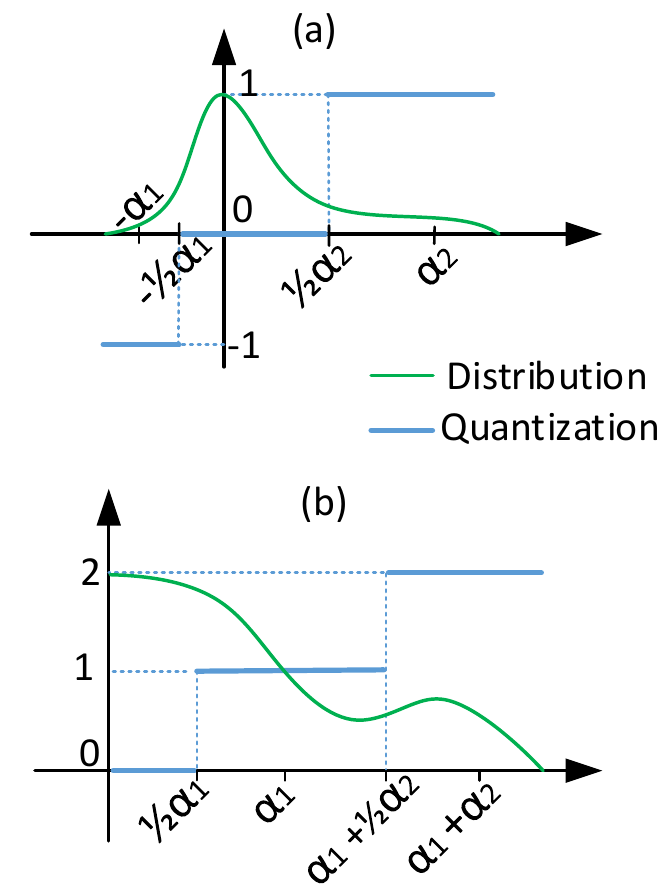}
\end{adjustbox}
\caption{The proposed non-uniform ternary quantization for (a) a tensor in the real domain; (b) a tensor which only contains non-negative values. The vertical axis represents the quantized domain and the horizontal axis denotes the real domain. The green curve indicates the distribution of the full-precision tensor and the blue line shows the quantized values by discretizing the full precision data according to the learned thresholds. We aim to learn the optimal step size of each quantization level.}
\label{fig:quantization}
\end{figure}

\begin{figure}[htb!]
\centering

\begin{subfigure}{.4\textwidth}
  \centering
  \includegraphics[width=.9\linewidth]{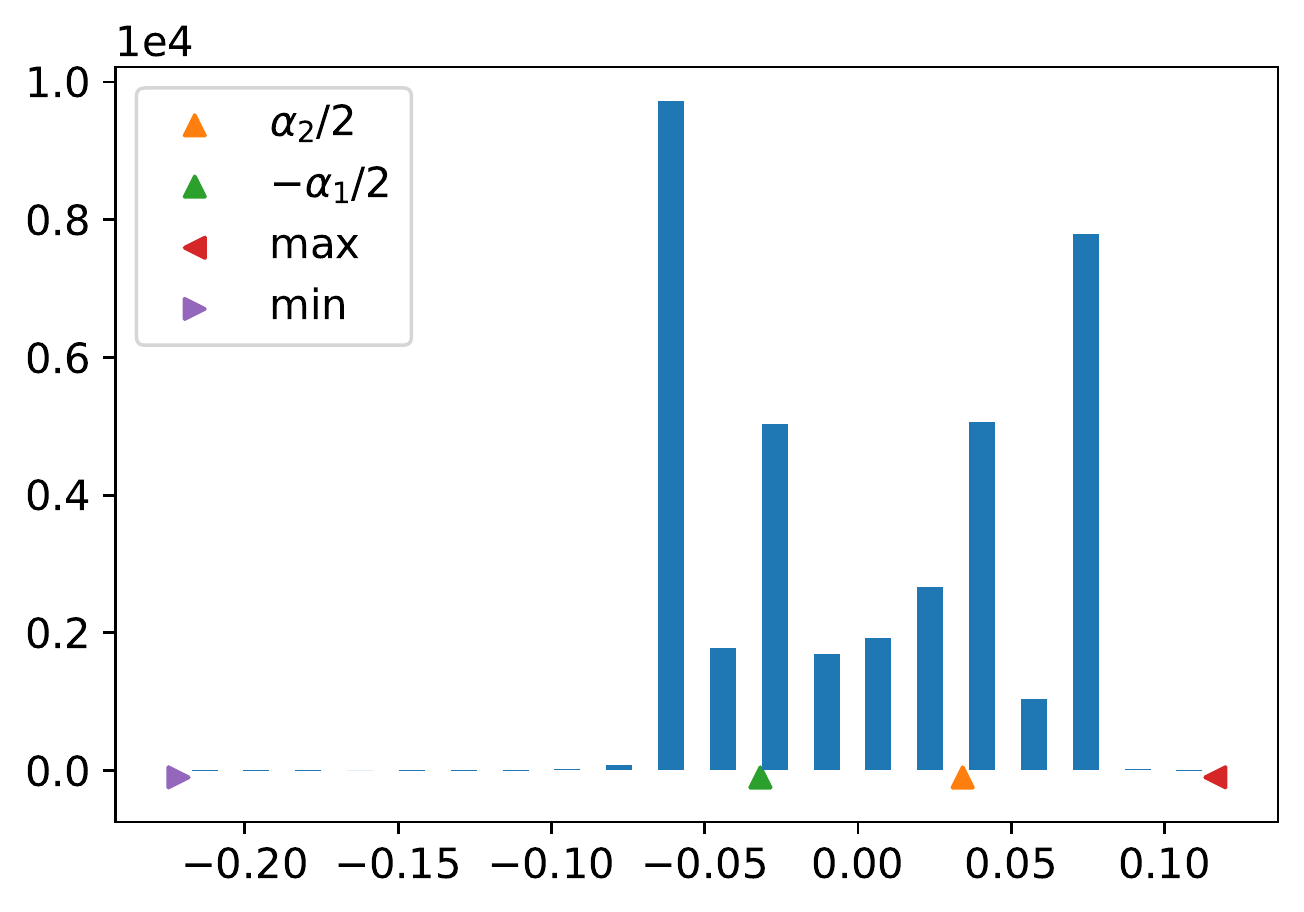}
  \caption{layer1\_1\_conv2}
\end{subfigure}
\begin{subfigure}{.4\textwidth}
  \centering
  \includegraphics[width=.9\linewidth]{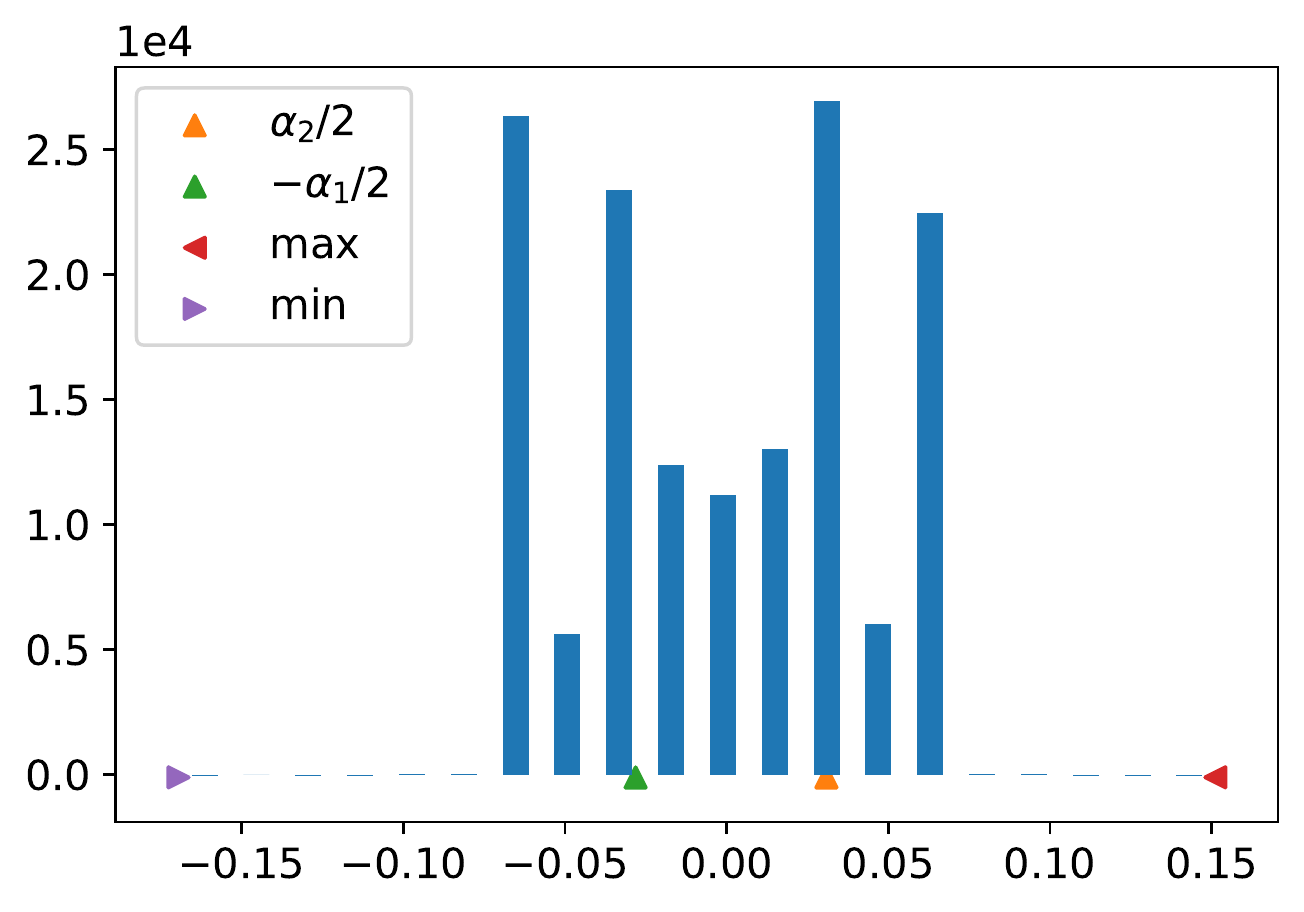}
  \caption{layer2\_1\_conv2}
\end{subfigure}
\begin{subfigure}{.4\textwidth}
  \centering
  \includegraphics[width=.9\linewidth]{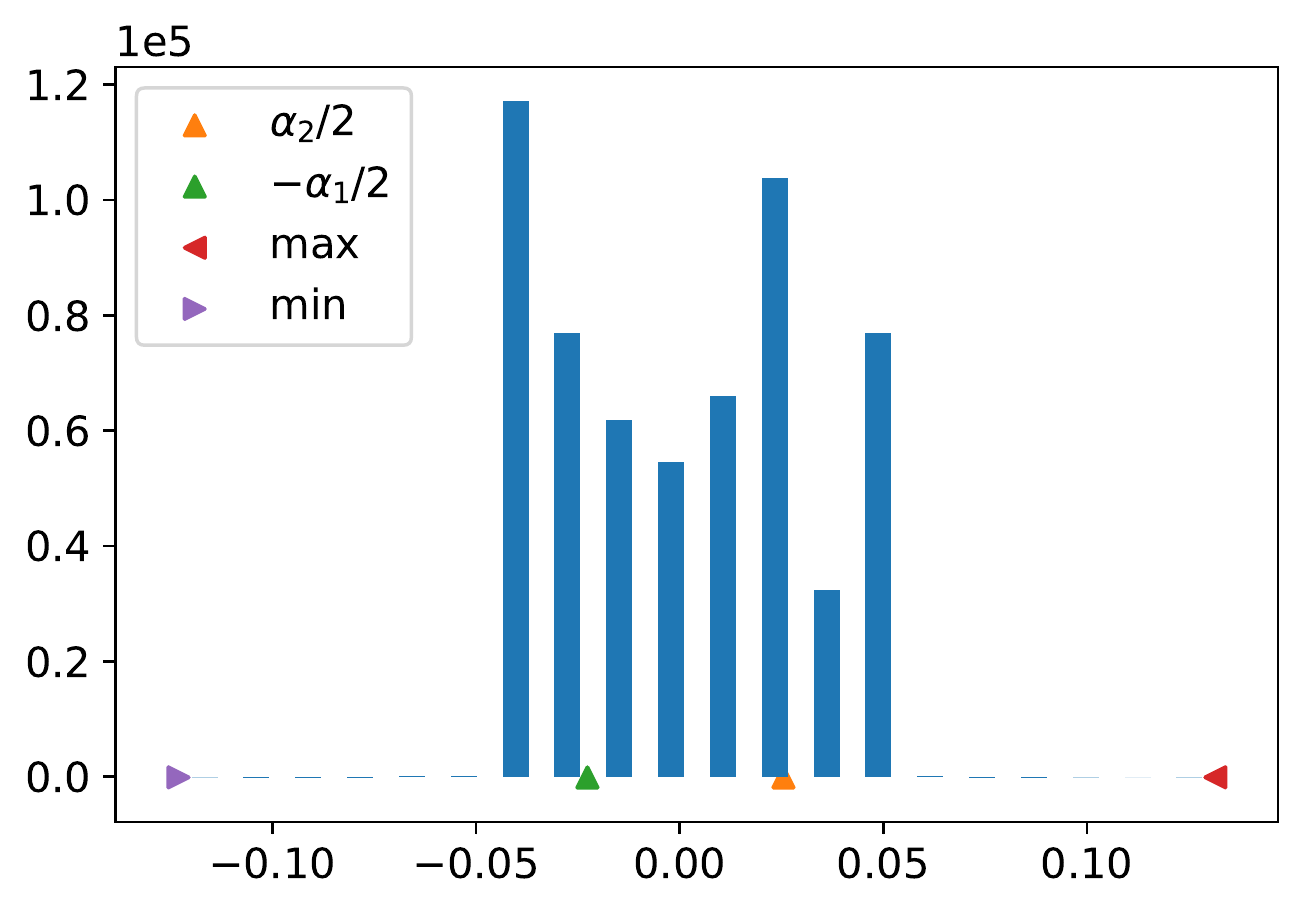}
  \caption{layer3\_1\_conv2}
\end{subfigure}
\begin{subfigure}{.4\textwidth}
  \centering
  \includegraphics[width=.9\linewidth]{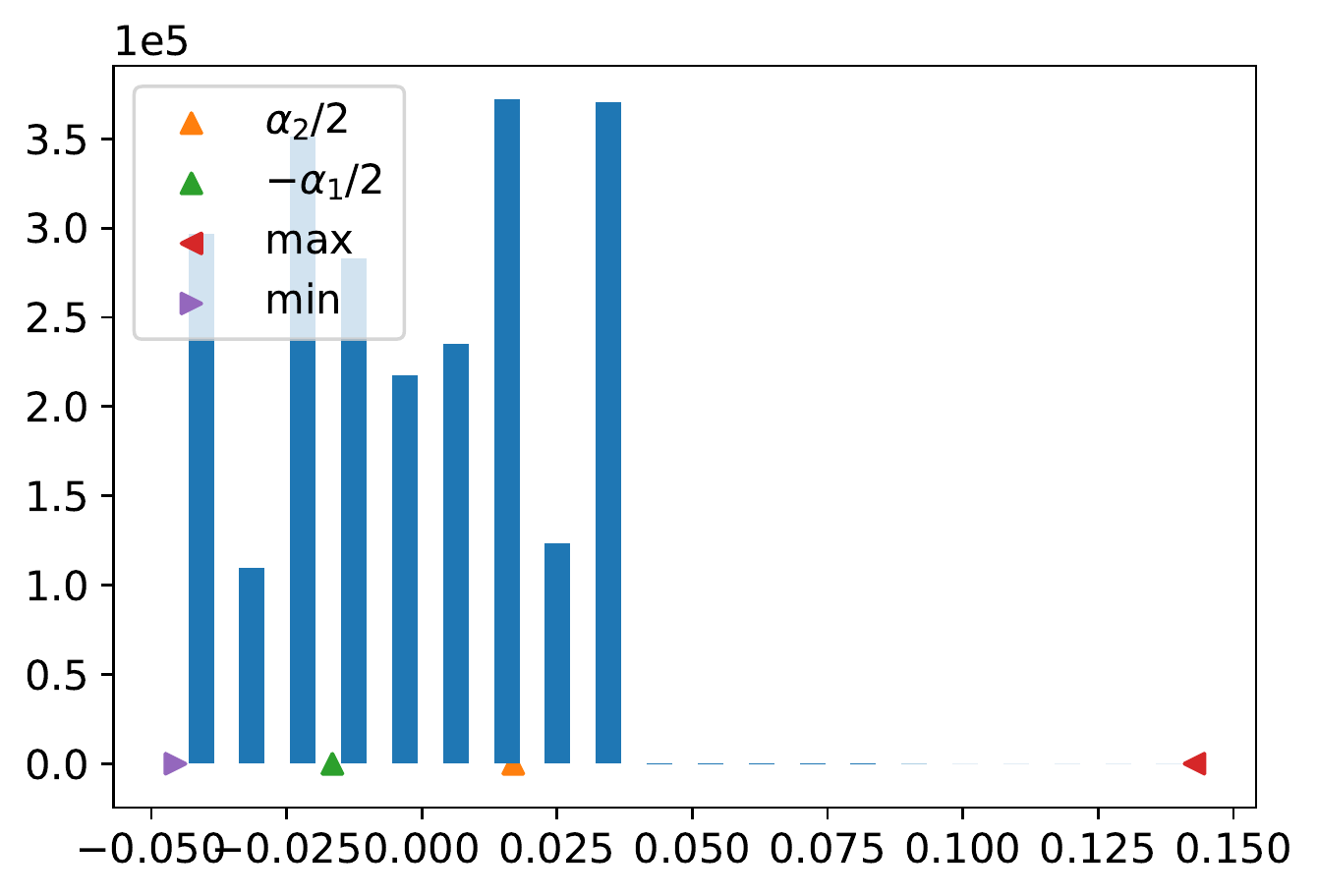}
  \caption{layer4\_1\_conv2}
\end{subfigure}

\caption{Weight quantization for ResNet-18.} \label{fig:weight-distribution}
\end{figure}

\begin{figure}[htb!]
\centering

\begin{subfigure}{.4\textwidth}
  \centering
  \includegraphics[width=.9\linewidth]{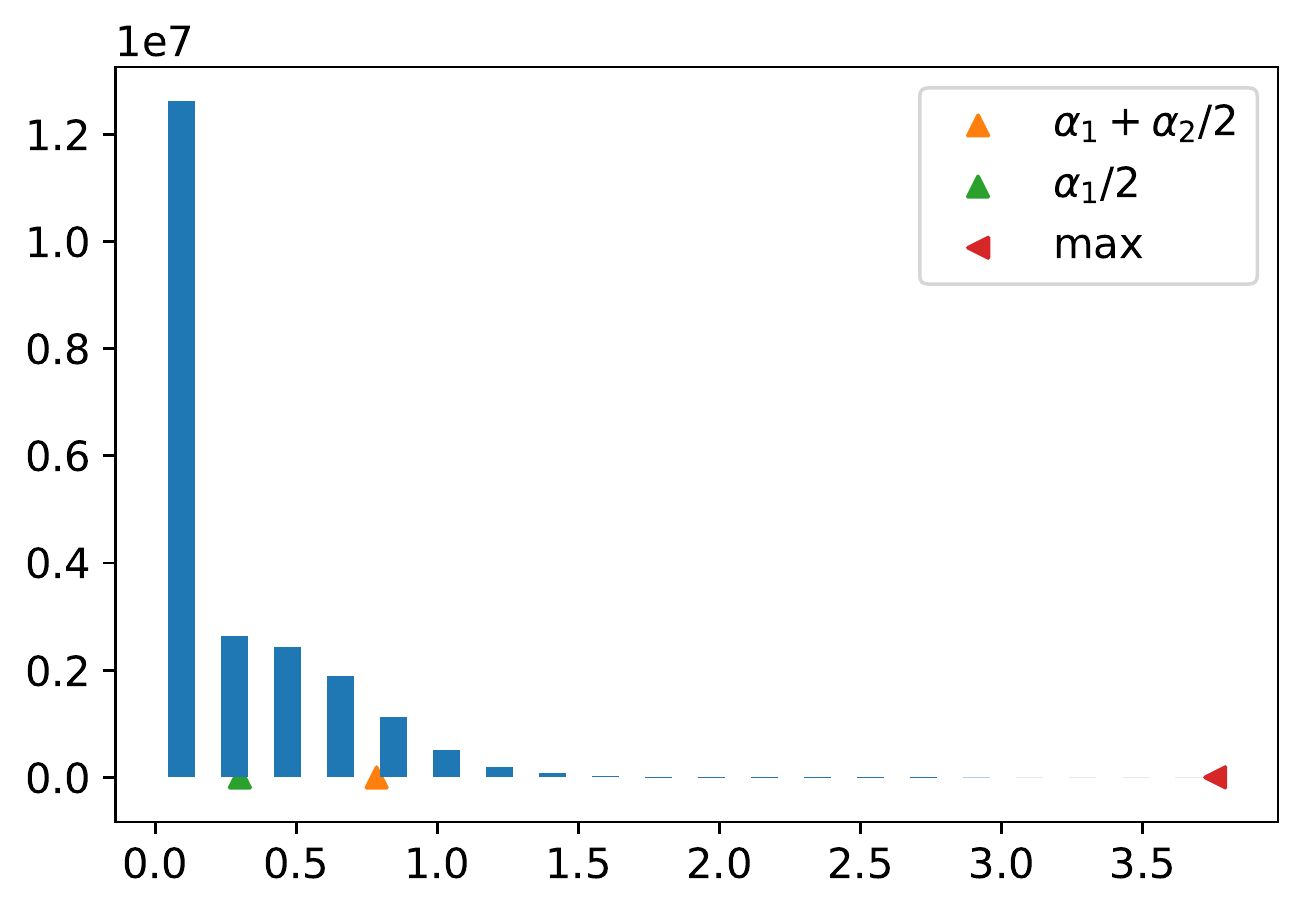}
  \caption{layer1\_1\_conv2}
\end{subfigure}
\begin{subfigure}{.4\textwidth}
  \centering
  \includegraphics[width=.9\linewidth]{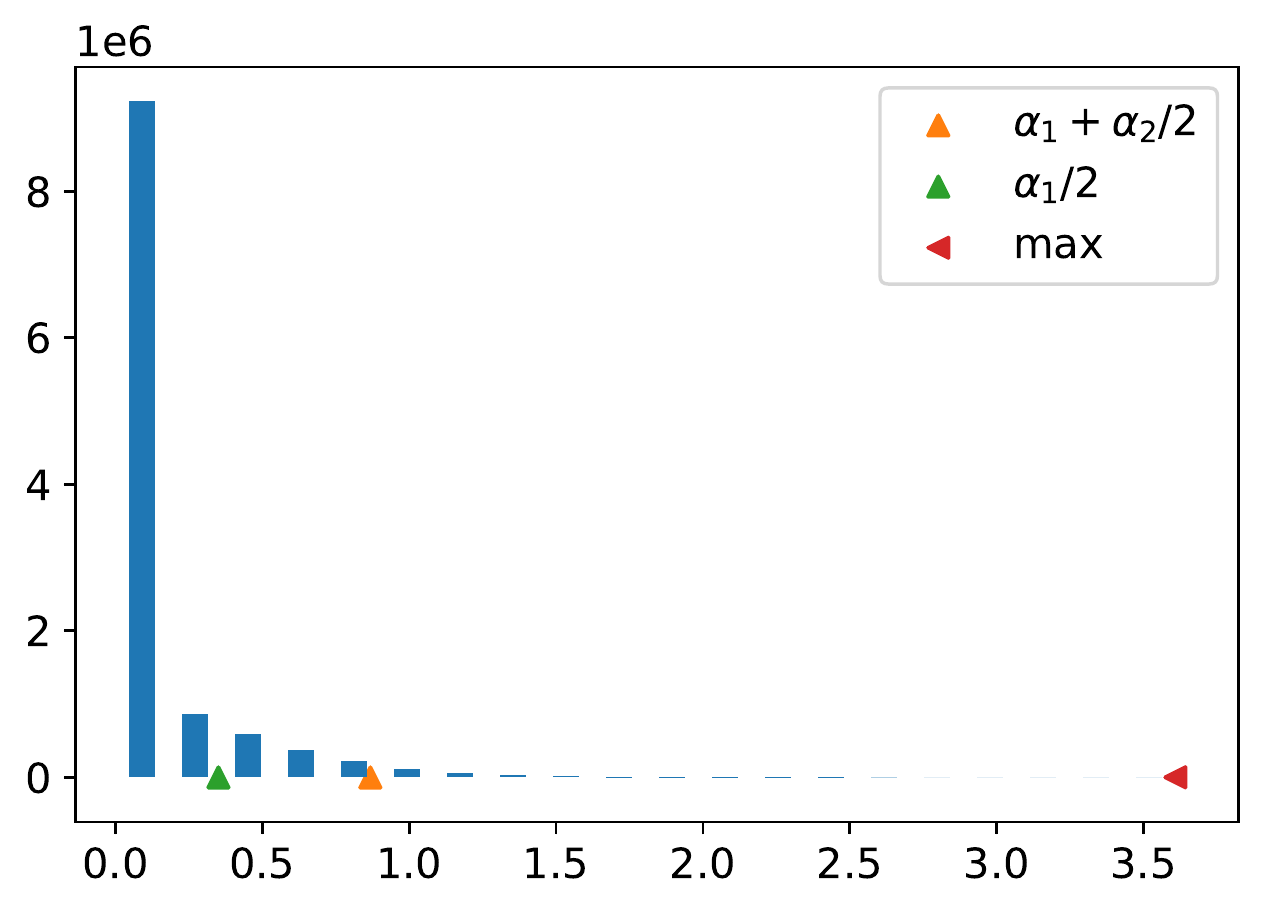}
  \caption{layer2\_1\_conv2}
\end{subfigure}
\begin{subfigure}{.4\textwidth}
  \centering
  \includegraphics[width=.9\linewidth]{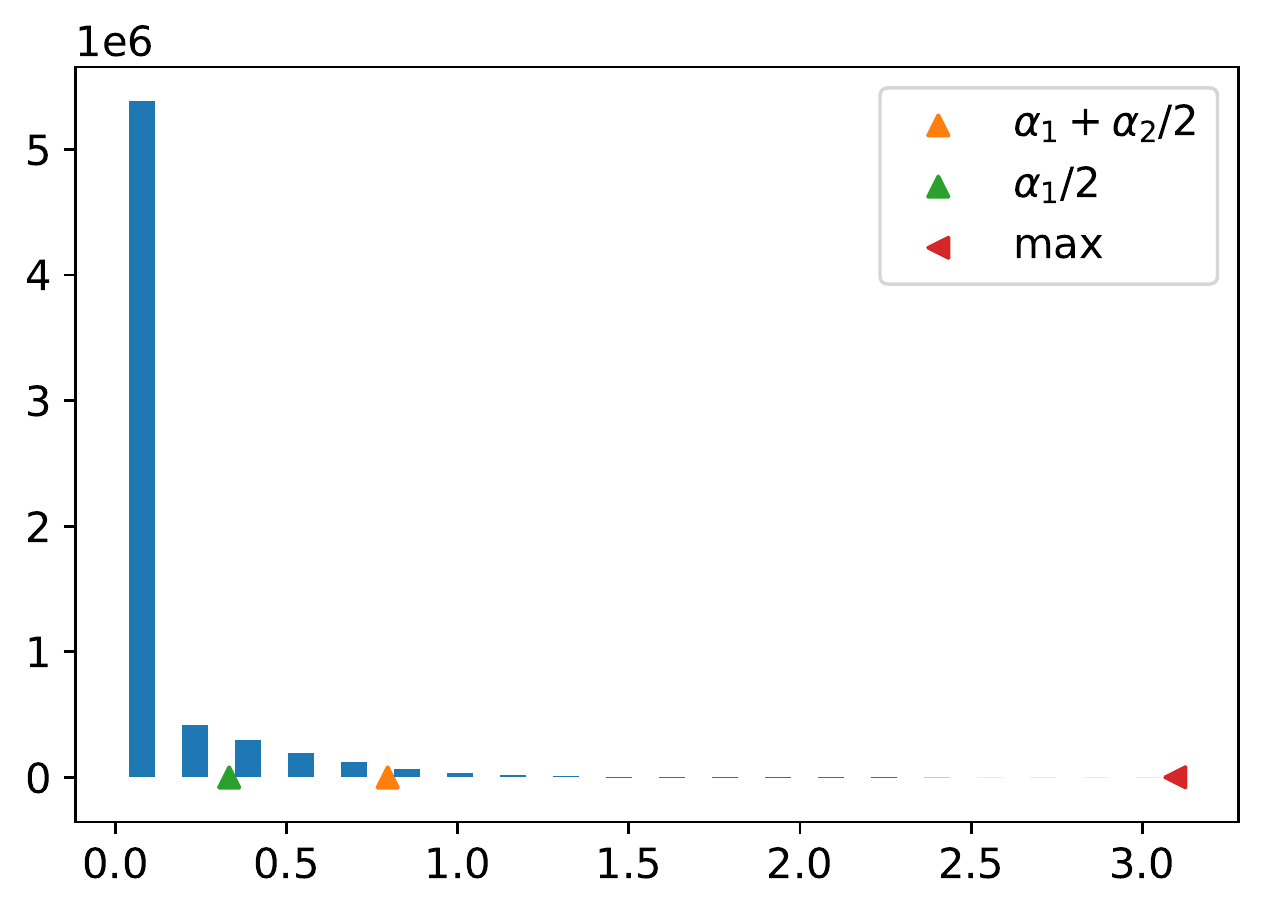}
  \caption{layer3\_1\_conv2}
\end{subfigure}
\begin{subfigure}{.4\textwidth}
  \centering
  \includegraphics[width=.9\linewidth]{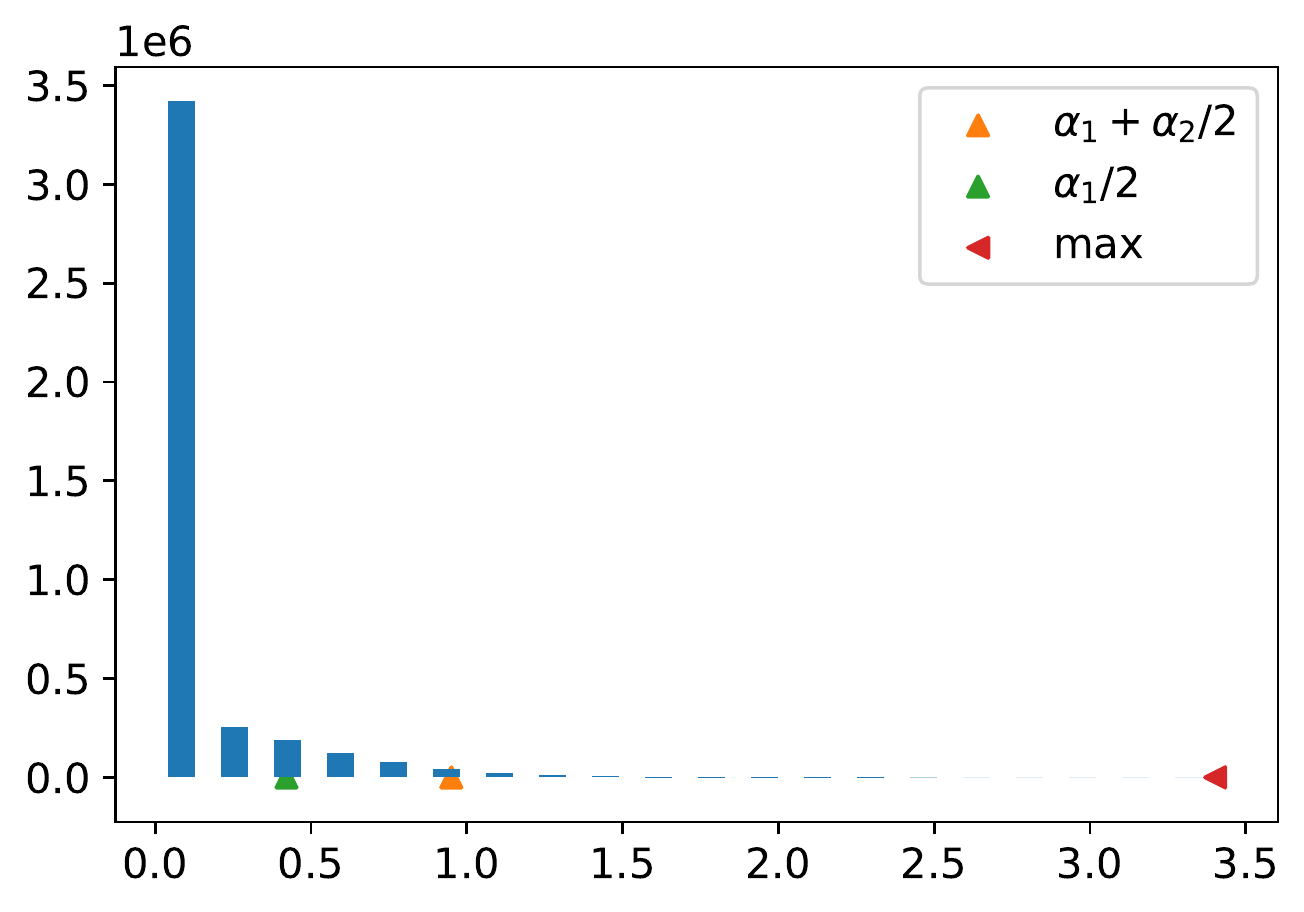}
  \caption{layer4\_1\_conv2}
\end{subfigure}

\caption{Activation quantization for ResNet-18.} \label{fig:activation-distribution}
\end{figure}

In this paper, we
discretize the full-precision tensor into the ternary quantized values with trainable non-uniform step sizes.
For each quantized layer, we learn two parameters $\alpha = \{\alpha_1, \alpha_2\}$ for weights and activations separately. As illustrated in Figure \ref{fig:quantization}, the quantization thresholds are directly related to the learnt parameters $\alpha$. When the two scale factors are identical ($\alpha_1 = \alpha_2$), the proposed quantization algorithm reduces to the uniform step size quantization \cite{esser2019learned, zhou2016dorefa, jung2019learning}. Thus, it is interesting to investigate the properties of the learnt $\alpha$. We plot the distribution of the full-precision weights and activations as well as the corresponding quantization thresholds on ResNet-18 in Figure \ref{fig:weight-distribution} and \ref{fig:activation-distribution}, respectively. We list the statistics of four layers in ResNet-18. Besides, the max value, min value and the two quantization thresholds of the tensors are marked in each sub-figure. On the one hand, from Figure \ref{fig:weight-distribution}, we observe that the distribution of the weight in the model varies a lot in different layers.
In order to reduce the information loss during quantization, we propose to learn the quantization thresholds automatically to better fit the data distribution. On the other hand, Figure \ref{fig:activation-distribution} demonstrates that, the full-precision activations consist of dense relative small values and sparse relative large values. The number of elements of each interval is unbalanced. It can be seen from Figure \ref{fig:activation-distribution} that the quantization step sizes learnt based on the stochastic gradient descent are non-uniform ones and differ at different layers.

\section{More Execution Time Benchmarks}

We present more acceleration benchmark results in this section.

First, for convenience of comparison, we list the the exact execution time for the layer-wise benchmark described in experiments section in Table \ref{tab:actual-time-1} and Table \ref{tab:actual-time-2}, respectively. We use Q821/Q835/2080Ti to indicate the Qualcomm 821, 835 and Nvidia 2080Ti separately. Besides, ``bin'' represents binary and ``ter'' means ternary in Table \ref{tab:actual-time-1} and Table \ref{tab:actual-time-2}.
We measure the layer-wise execution time for convolution layers (kernel size = $3 \times 3$, padding = 1, stride = 1, the same number of input and output channels and batch size = 1) with six different shape configurations.
For the first four cases, we fix the channel number to be 64 and increase the resolution from 28 to 224. For the last two cases, we fix the resolution to be 56 and double the channel number from 64 to 256.

\begin{table*}[ht!]
\centering
\caption{Exact execution time ($\rm ms$) on embedded-side platforms. We run 5 times and report the mean results.}
\label{tab:actual-time-1}
\begin{tabular}{r | c | r r r r r r}
 Device & A/W & case1 & case2 & case3 & case4 & case5 & case6  \\
\shline
\multirow{3}{*}{Q821} & bin/bin & 0.6 & 1.2 & 2.6 & 8.3 & 2.3 & 6 \\
 & ter/ter & 0.9 & 1.7 & 4.4 & 14.8 & 4.2 & 13.5  \\
 & 2/2 & 1.9 & 3.5 & 8 & 24.9 & 7.2 & 21.2  \\
\hline
\multirow{3}{*} {Q835} & bin/bin & 0.7 & 0.9 & 1.9 & 6.1 & 1.8 & 4.8 \\
 & ter/ter & 0.9 & 1.5 & 3.7 & 13.2 & 3.2 & 12.7 \\
 & 2/2 & 2.1 & 3 & 6.5 & 20.5 & 5.6 & 15.8  \\
\end{tabular}
\end{table*}

\begin{table*}[ht!]
\centering
\caption{Exact execution time ($\mu s$) on server-side platforms. We run 5 times and report the mean results.}
\label{tab:actual-time-2}
\begin{tabular}{r | c | r r r r r r}
 Device & A/W & case1 & case2 & case3 & case4 & case5 & case6 \\
\shline
\multirow{3}{*} {2080Ti}
 & bin/bin & 11 & 12.5 & 19.5 & 56.5 & 22 & 55.5 \\
 & ter/ter & 11 & 14.5 & 26 & 90 & 34 & 87 \\
 & 2/2 & 38 & 44 & 73.5 & 274 & 91 & 217.5  \\
\end{tabular}
\end{table*}

\section{Evaluation on CIFAR-10}

\subsection{Training hyper-parameters on CIFAR-10}

We follow the hyper-parameter setting in previous works \cite{zhang2018lq, 2018peisongTSQ} to train the networks on the classification task. Specifically, for ResNet-20 on CIFAR-10, we train up to 200 epochs. The initial learning rate starts from 0.1 and is divided by 10 at epoch of 82 and 123, respectively. We use a weight decay of 1e-4 and a batch size of 128. For VGG-Small on CIFAR-10, the learning rate begins with 0.02 and is divided by 10 at epoch of 80 and 160, separately. The weight decay is set to be 5e-4, batch size to be 128 and total epochs to be 200. For NIN on CIFAR-10, we train 90 epochs with the initial learning rate to be 1e-2. The learning rate is divided by 10 at epoch of 30 and 60. Weight decay is set to be 1e-5.

\subsection{Comparison on CIFAR-10}

We further demonstrate the effectiveness of the proposed method on CIFAR-10 \cite{krizhevsky2009learning} dataset. For CIFAR-10, the input images are firstly padded with 4 pixels and then cropped into 32$\times$32 samplings. Random horizontal flip is employed for data augmentation. NIN \cite{Lin2014NetworkIN}, VGG-Small \cite{simonyan2014very} and ResNet-20 are employed for the evaluation.
Comparisons between our FATNN and other quantization algorithms are listed in Table \ref{tab:cifar-10}.

\begin{table}[bt!]
\centering
\caption{Top-1 accuracy (\%) comparisons between our FATNN and other algorithms, including RTN \cite{Li2020RTN}, HWGQ \cite{Cai_2017_CVPR}, LQ-Net \cite{zhang2018lq} and TSQ \cite{2018peisongTSQ} on CIFAR-10 dataset.}
\label{tab:cifar-10}
\scalebox{0.90}{
\begin{tabular}{l | c | c | c | c}
Method & A/W & ResNet-20 & NIN & VGG-Small\\
\shline
   & 32/32 & 92.1 & 89.8 & 93.8 \\
\hline
 FATNN & ter/ter & \bf{90.2} & \bf{89.9} & \bf{93.7}\\
   RTN & ter/ter & - & 89.6 & -\\
HWGQ & 2/1 & - & - & 92.5 \\
 LQ-Net & 2/1 & 88.4 & - & 93.4\\
 TSQ & 2/ter & - & - & 93.5 \\
 LQ-Net & 2/2 & 90.2 & - & 93.5\\

\end{tabular}}
\end{table}

From Table \ref{tab:cifar-10}, we observe that for ResNet-20 and VGG-Small, the proposed method is able to achieve comparable accuracy for ternary networks compared with higher bit counterparts (``2/2'' or ``2/ter'') quantized by LQ-Net and TSQ. Moreover, it surpasses RTN, HWGQ and LQ-Net on all the corresponding ``2/1'' or ``ter/ter'' networks. For NIN, our FATNN even beats the full-precision counterpart.
It implies that network quantization has certain kind of regularization effect to suppress the over-fitting problem. Overall, the proposed FATNN demonstrates superior performance on the CIFAR-10 dataset.

\end{document}